\documentclass[10pt,onecolumn]{IEEEtran}
\usepackage{geometry}
\usepackage[switch]{lineno}
\usepackage[cmex10]{amsmath}
\usepackage{amssymb,latexsym,amsfonts,stmaryrd}
\usepackage{amsthm}
\usepackage{mathtools}
\usepackage{graphicx}
\usepackage[noadjust]{cite}
\usepackage{algorithm}
\usepackage{algorithmic}
\usepackage{color}
\usepackage{soul}
\usepackage{lineno}
\usepackage{mathptmx}
\usepackage[ruled,vlined,algo2e]{algorithm2e}
\usepackage{longtable}
\usepackage{etoolbox}
\usepackage{adjustbox}
\usepackage{multirow}
\usepackage{subcaption}
\makeatletter
\patchcmd{\@makecaption}
  {\scshape}
  {}
  {}
  {}
\makeatother

\usepackage{array}% http://ctan.org/pkg/array
\usepackage[hyphens]{url}
\usepackage[hidelinks]{hyperref}
\hypersetup{breaklinks=true}
\usepackage{cite}

\usepackage{natbib}
\usepackage{cleveref}

\setcitestyle{authoryear}
 
\title{A deep real options policy for sequential service region design and timing}

\author{
  \IEEEauthorblockN{Srushti Rath and  Joseph Y.J. Chow \\}
  \IEEEauthorblockA{C2SMART University Transportation Center}\\
    Department of Civil and Urban Engineering\\
    New York University, NY, USA  \\
    Email: srushti.rath@nyu.edu,  joseph.chow@nyu.edu}
\begin{document}
\maketitle
\section*{Abstract}
As various city agencies and mobility operators around the world navigate toward innovative mobility solutions, there is a need for strategic flexibility in well-timed investment decisions in the design and timing of mobility service regions, i.e. cast as "real options". This problem becomes increasingly challenging considering the multiple interacting real options in such investments. We propose a scalable machine learning (ML) based real options (RO) framework for multi-period sequential service region design and timing problem for mobility-on-demand (MoD) services, framed as a Markov
decision process with non-stationary stochastic variables. A value function approximation policy from the literature uses multi-option least squares Monte Carlo simulation to get a policy value for a set of interdependent investment decisions as deferral options (\emph{i.e.}, $CR$ policy). The objective is to determine the optimal selection and timing of a set of zones to include in a service region (which to add now, which to defer to a later time to reconsider). However, prior work required explicit enumeration of all possible sequences of investments (\emph{i.e.} 5 options translate to 120 sequences). To address the combinatorial complexity arising from sequence enumeration, we propose a new variant "deep" real options policy using an efficient recurrent neural network (RNN) based ML method (\emph{i.e.}, $CR-RNN$ policy) to sample sequences to forego the need for enumeration, making the network design and timing policy tractable for large scale implementation. Experiments based on multiple service region scenarios in New York City demonstrate the efficacy of the proposed policy in substantially reducing the overall computational cost (\emph{i.e.}, time reduction associated with the RO evaluation of more than 90\% of total investment sequences is
achieved), with zero to near-zero gap compared to the benchmark. We validate the model in a case study of sequential service region design for expansion of MoD services in Brooklyn, NYC, under service demand uncertainty. Results show that using the CR-RNN policy in determining optimal real options investment strategy yields a
similar performance ($\approx$ 0.5\% within the CR policy value) with significantly reduced computation time (about 5.4 times faster).\\

%To illustrate the efficacy of the proposed method, we conduct various experiments on multiple service region scenarios in New York City (NYC), where our findings show a significant reduction in the total computational cost by using the RNN model \
%(\emph{i.e.}, reduction associated with evaluation of more than 90\% investment sequences with less than 1\% prediction gap).

\noindent Keywords:  Sequential service region design, Real options policy, Recurrent neural networks, Approximate dynamic programming, Flexible investment, Markov decision process, Demand uncertainty\\

%% INTRODUCTION %%
\section{Introduction} \label{sec:introduction}
\begin{figure}[!htb]
\begin{center}
\includegraphics[scale=.48]{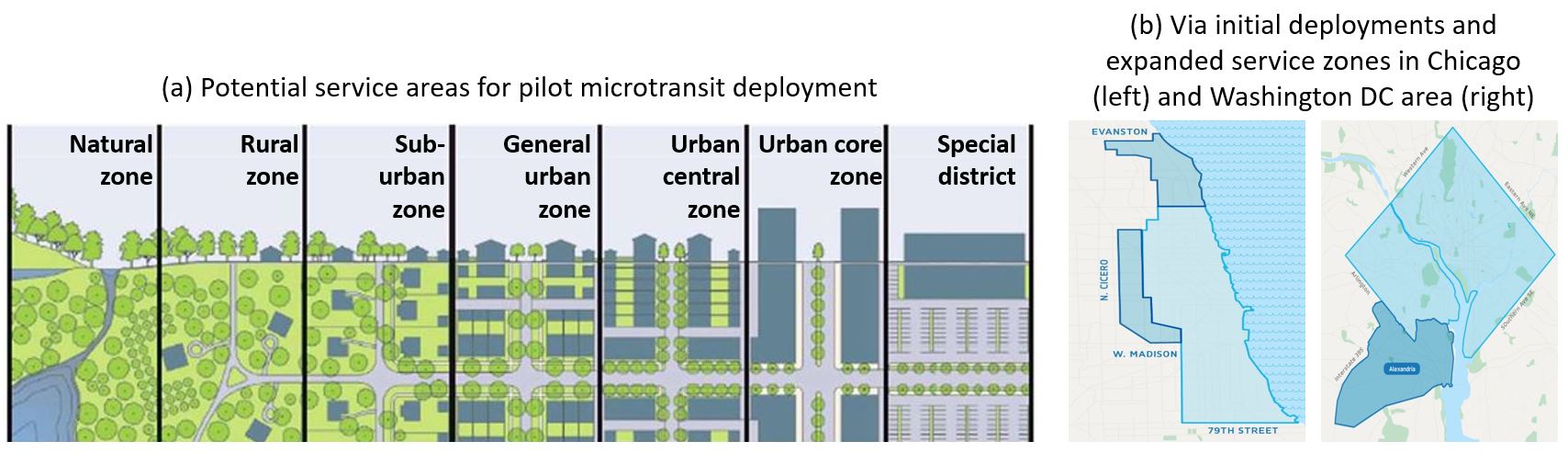}
\caption{(a) Examples of potential service areas in a city for pilot microtransit deployment (\citealp{sumc}), (b) Via (microtransit) service zones in Chicago and greater Washington DC area, U.S. (ridewithvia.com/news/); initial deployments and expanded areas are highlighted in lighter and darker shades respectively. } 
\label{fig:serviceareas_sumc} 
\end{center}
\end{figure}

Despite the increasing popularity and growth of emerging transportation services in the broad mobility marketplace (such as ridesharing, MoD, mobility-as-a-service), there are inevitable uncertainties associated with them. Such uncertainties include changing customer preferences, travel needs, fuel costs, regulations, environment and other factors. Planning the deployment, operation, and expansion of such services to multiple cities involves significant operation costs, and requires capital investments from mobility operators and public agencies. An important prerequisite for the strategic expansion of such services to new markets (\emph{e.g.,} initial service deployment to a specific area in a city as shown in \Cref{fig:serviceareas_sumc}(a)), or growing the existing service coverage (\emph{e.g.,} expansion to new zones as shown in \Cref{fig:serviceareas_sumc}(b)) is the cost and benefit assessment of the investment decisions related to service region design.

Service region design is a non-trivial problem. Selecting too large of a coverage area reduces the response time of a given fleet; too small of an area limits the potential demand for the service. Selecting a highly dense area for service may subject the fleet to unwanted congestion from local traffic; areas with sparse populations risk having low ridership. In the design of such service regions, due to increasing public private partnerships between city agencies and transportation network companies (\citealp{shaheen2020mobility}), optimizing the investment timing is a challenging problem with significant long-term impact. The challenges primarily stem from the need to account for time-dependent uncertainties while being constrained by funding and budget requirements.

Identifying service areas is a key aspect of the project design phase; this is needed for planning allocation of resources and funds both from public agencies and mobility operators (\citealp{sumc}).
However, service area identification is not static, in the sense that future expansion may be needed on top of the initial service area. In an environment evolving with uncertainty, timed investment decisions face additional complexities while accounting for such future expansion.
Investing in a new transportation service or expanding the same to specific zones too early may yield low ridership and revenue shortfalls, while waiting for a longer time period may result in large social costs and reduced benefits. Furthermore, the volatility in the uncertain elements may vary based on multiple factors including the community or city specific socio-demographic and market conditions.
Many prior studies have emphasized the need to incorporate flexibility in strategic decision-making in transportation to ensure performance enhancing design in response to uncertain elements (\citealp{ cardin, chow2011network}).

Consider, for example, the microtransit company Via, which has substantial deployments in more than 15 countries around the world in partnerships with various transit agencies. In 2016, Via started their on-demand transit services in Washington DC and later expanded to nearby areas in Arlington, followed by Alexendria in Virginia in 2019 (\Cref{fig:serviceareas_sumc}(b)). However, the company had to shut down its operations in Washington DC (\citealp{masstransit}) due to reduced benefits during the COVID-19 pandemic. Around the same time, Via expanded their services in West Sacramento, California from two to nine (operator defined) zones (\citealp{westerncity}), while in Arlington, Texas, their service coverage was increased from 41\% area to the remaining parts of the city, owing to improved ridership (\citealp{keranews}). Therefore, it would be more useful for decision makers to have quantitative tools to incorporate future uncertainties associated with the inherent service adoption of such technologies and external market factors pertaining to different areas. 
Such tools can support large-scale flexible investment decisions, where flexibility may include actions such as the option to invest or defer the project at different stages of the investment horizon. This can specifically help mobility operators and planners evaluate and compare the benefits of investing in select service areas in multiple cities, to identify areas of investment now under a given budget (\emph{e.g.}, intial deployments in some cities and service expansion across other cities) while considering long-term uncertainties in different locations. 
In other words, when decisions can be made continuously over time and the underlying conditions may also evolve over time, the value of decision flexibility needs to be accounted for to identify the best long term strategy.

\begin{figure}[!htb]
\begin{center}
\includegraphics[scale=.45]{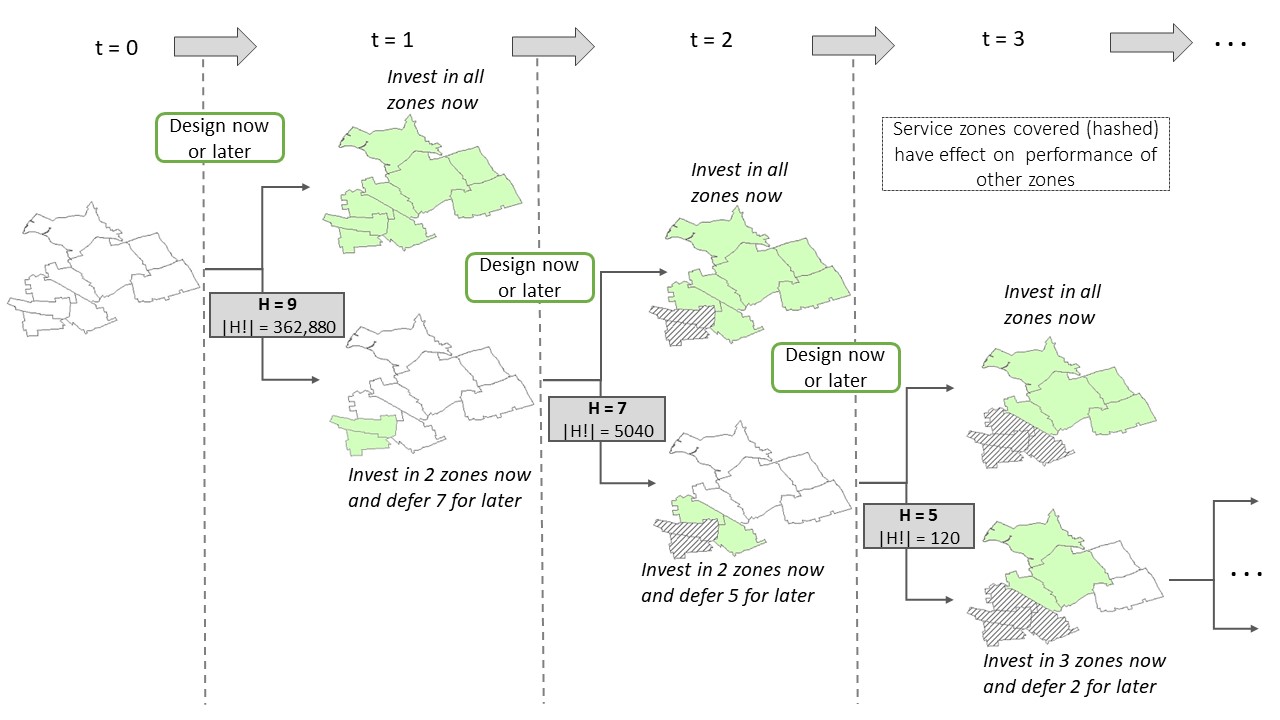}
\caption{Illustration of sequential service region design and timing problem (for a set of $\mathcal{H}$ zones); green highlighted service zones denote investment decisions in those zones (\emph{i.e.}, invest immediately) at time $t$, hashed zones represent zones already designed (invested) in previous time steps, and no highlights at time $t$ represent zones where it is optimal to defer the investment to a later time period.} 
\label{fig:illustrative_example} 
\end{center}
\end{figure}

Traditional cost-benefit analyses based on the net present value (NPV) do not incorporate flexible investment actions in response to uncertainties, since they tend to ignore the potential value of investment that arises from deferring the option to wait for more information in an uncertain environment. Consider a set $\mathcal{H}$ of geographically contiguous zones ($|\mathcal{H}| = H$) constituting a potential service region for MoD services as a portfolio of interdependent projects or assets (as illustrated in \Cref{fig:illustrative_example}); for each zone, is it optimal to invest in serving the zone immediately or postpone that zone to a later time (within a time horizon) or avoid completely to maximize the return on the portfolio under demand uncertainty? To answer this set of questions, the broad concept of real options can be applied (\citealp{trigeorgis}). Real options theory is an investment evaluation method prominent in corporate finance; it allows decision makers the freedom to make the best choices as to when and where to make a specific capital investment among multiple financial options. This perspective considers the value to react flexibly to future uncertainty as opposed to the conventional investment rules (based on NPV), and captures the impact of varying uncertainty levels on investment decisions. Based on the real options principle, the NPV is expanded upon by considering the \textit{option premium} (\Cref{eq:expanded_NPV}); this incorporates the value due to options which would not be captured by a static NPV analysis. 
\begin{align}
    \text{Expanded NPV = base NPV} + option \; premium \label{eq:expanded_NPV}
\end{align}

For multiple interacting options (\emph{i.e.}, compound options in real options theory), say in timing decisions of multiple investments within a network, the premium is further divided into a deferral premium and a network redesign premium (\citealp{chow2011network}), and the investment payoff of a zone (project) $h \in \mathcal{H}$ depends on the prior investment decisions (\citealp{gamba2003}). Thus, for each investment candidate project $h$, what has been invested before and what is to be invested at subsequent time
steps can be defined in terms of an investment sequence as a well-defined lower bound on the option value that is missing the redesign premium (\citealp{chow2011network}). For convenience, we call this lower bound RO-based policy the "Chow-Regan (CR) policy" (\citealp{chow2016reference}). The total number of possible sequences for a portfolio of $H$ compounded projects is $H!$, and the CR policy selects the sequence that offers the highest initial option value (\citealp{chow2011network}). However, when the number of projects ($H$) increases, the possible number of sequences ($H!$) increases drastically, hence, evaluating the CR policy value for each of the enumerated sequences becomes computationally expensive (and even intractable) as the value of $H$ increases, and has restricted its applicability to scenarios with timing small sets of options (say, not more than 6 to 9 options). 

This task becomes non-trivial in multi-period look-ahead sequential service region design where at each time period ($t$), the idea is to optimize the timing of a set of investments in order find the optimal policy (from $H!$) under a dynamic setting. For example, \Cref{fig:illustrative_example} illustrates a sequential service region design and timing problem for MoD services (\emph{e.g.,} microtransit), where it is assumed that the user demand for the service is uncertain and evolves over time following a stochastic process. Now, traditional NPV may suggest investing in all $9$ zones at $t=1$. However, as per a real options based policy, it would be better to invest in only 2 zones (highlighted out of $H = 9$ zones) and defer the investment of $7$ zones until a later time (\emph{i.e.,} until a finite horizon, at which point further deferral indicates rejection), since this has more value in terms of the expanded NPV. This optimal policy is obtained by evaluating $9!$ possible investment sequences (for $t=1$). Considering the investment in these $2$ zones (at $t=1$), the decision process is repeated for the next time period (with $H = 7$ zones and $7!$ sequences), and so on in a rolling (finite) horizon, allowing the service region design to adapt to new conditions. In this particular example (which illustrates one projected realization of future demand uncertainty), just for $t=1$, 2 and 3, one would require to evaluate about $368,040$ investment sequences for the optimal policies, and to conduct such calculations over multiple simulations (demand realizations) with sequence enumeration is computationally very expensive. This inevitably calls for efficient solution methods with lower computational complexity so as to be implementable in practice. In fact, this is not just restricted to service region design, but applies to various other sequential network design problems in transportation. To the best of our knowledge, no prior studies deal with such combinatorial complexity arising in network design and timing decisions for a large number of projects. 

Therefore, to address this challenge, we propose a new ML based method to reinforce the CR policy to solve the real options based sequential service region design and timing problem for MoD services. More precisely, instead of evaluating all $H!$ investment sequences for $H$ candidate zones in a region as in the conventional \textit{CR} policy, we train a recurrent neural network (RNN) based sequence classifier with a small (randomly sampled) fraction of sequences, and use the trained RNN to predict top sequences (\emph{i.e.}, sequences with high policy values) from the large set of remaining candidate sequences. The RNN training involves the novel use of a gap relative to a reference policy value \citealp{chow2016reference}, which can account for the uncertainty in a setting that other reference measures like "competitive ratio" \citealp{karp1992line} cannot. In this sense, the RNN is used as a heuristic optimization technique to obtain the CR policy: let us call this the \textit{CR-RNN} policy. This should avoid the need to perform computationally expensive RO based calculations on all $H!$ sequences, to obtain significant time savings. The motivation behind such a data-driven approach leveraging RNNs stems from the intuition that RNNs can learn patterns in the investment sequences which lead to high policy values (from a small training data set), and such predictive capability can then be used to evaluate a large set of (remaining) test sequences. We show that sequence evaluation can be effectively framed as a binary classification task, where the RNN's objective is to just predict whether a given sequence is close enough (\emph{i.e.}, within a certain range) to the optimal policy value across all sequences.
This novel methodology leverages the power of neural networks towards learning patterns in sequences, and opens up new possibilities at the intersection of sequential network design and deep learning models. 

Our main contributions can be listed as follows: 
\begin{enumerate}
    \item We adapt the CR policy (\citealp{chow2011network,chow2016reference}) as a value function approximation method in approximate dynamic programming (\citealp{powell2007approximate}) to solve the lower bound sequential service region design and timing problem for MoD services, framed as a Markov decision process with non-stationary stochastic variables. The objective is to determine the optimal investment strategy for the service zones in a region \emph{i.e.}, investment or deferral decisions (similar to \cite{chow2011network}'s treatment of real option value in network design models). We model zone interactions in terms of interrelated stochastic origin-destination demand across zones, and consider heterogeneous volatility in the stochastic elements. This can support planners towards optimal dynamic decision-making over time in design of service regions under demand uncertainty for MoD services. Leveraging this approach, planners can avoid losses under unfavorable conditions and seize opportunities when beneficial.

    \item To address the high computational cost associated with evaluating all candidate investment sequences via the CR policy, we propose an RNN-based sequence classifier to reinforce the algorithm and avoid full sequence evaluation (CR-RNN policy), by making novel use of a reference policy to quantify the gap in a sequence option value. The classifier requires policy values to be computed only for a fraction of sequences to serve as training data. For the remaining fraction, the trained RNN is used to efficiently identify top sequences (with high policy values) to guide the sequential investment decisions. 
    We demonstrate the efficacy of our proposed method on different service region scenarios in NYC (with 7 and 8 zones where the CR policy can be compared against) along with sensitivity analysis on various parameters. We achieve a massive reduction in computation time (eliminating the need to evaluate about more than 90\% candidate investment sequences across different scenarios) using our proposed CR-RNN policy approach compared to the benchmark CR policy with significantly high prediction accuracy (\emph{i.e.}, leading to zero or near-zero gap with the CR policy value depending on the amount of data used for training the RNN classifier).
    
    \item In dynamically uncertain contexts, both public agencies and private mobility operators are confronted with decisions regarding the timing of their resource allocation and service expansion to different areas. %specifically while managing a portfolio of multiple cities. 
    We validate the model in a case study on NYC to demonstrate how our proposed method can be used to build effective decision support tools in this context, and share valuable insights. Because the cost, benefits, and volatility in evolution of demand of such mobility services may vary per city or city types, our research can benefit planners in efficiently evaluating the impacts of new services and policies by assessing trade-offs between present benefits and deferral option values for identifying effective resource allocation strategies.
\end{enumerate}

Additionally, if properly trained (with sufficient number of simulated scenarios), the proposed method can be be used to develop a generalized RNN based model that can be applied to varying region sizes.  Essentially, the proposed method can be used to efficiently valuate different scenarios for faster training of
the generalized ML model with varying region sizes (using zero padding, Piao and Breslin, 2018) along with other
region specific input parameters (including population characteristics, service volatility, and operation strategies). This can be built into a decision support tool which could benefit both public
agencies and mobility operators in the funding and resource allocation planning, as part of a complex cost-benefit
analyses process.

The paper is organized as follows. In Section~\ref{sec:related_work}, we provide a
brief literature review. Section~\ref{sec:methodology} explains the proposed method and the algorithmic techniques that we employ in our study; this is followed by experimental design in \cref{sec:experiments}, with results and discussion in Section~\ref{sec:results}. We conclude in Section~\ref{sec:conclusion}.\\

%%%% RELATED WORK %%%%%
\section{Related Work} \label{sec:related_work}
\subsection{Service region design}
The design of a service region may be examined in two ways. If the region is defined continuously, the problem is a boundary or cordon design problem. Cordon design has been studied in the context of congestion pricing (\citealp{zhang2004optimal}). If the region is divided into discrete zones, the problem involves selecting a contiguous set of zones such that a service delivery objective is optimized. For example, \cite{huang2018designing} call this the block design problem in city logistics.
Another approach is the use of connected subgraph problem (see \citealp{grotschel1990integer}) to select a set of zones with connectivity constraints with each other. However, these present a deterministic approach to designing the service regions and lack the time dimension in the decision-making process \emph{i.e.}, optimal timing in investing or designing a set of zones that may introduce additional benefits.
%In graph theory, the connected subgraph problem (see \citealp{grotschel1990integer}) is used often in telecommunications to find a subset of nodes such that every node is directly connected to at least one other node ($k=1$). When $k\geq2$, the problem is NP-hard (\citealp{garey1979computers}). 
%It can be used in network design to select a set of zones with connectivity constraints with each other. 
Studies focusing on the design of service regions for emerging mobility systems are limited. \cite{he2017} develop an optimization framework to design service regions for one-way electric vehicle sharing system. \cite{qi2018shared} use analytical models to determine optimal size of service zones for investigating adoption of shared mobility for last-mile delivery. \cite{zhou2021surrogate} explore an activity-based connected subgraph problem for designing service regions for shared automated vehicle fleets using MATSim simulation-based optimization. %\cite{he2017} for electric shared mobility,  \cite{qi2018shared} for last mile deliveries,
\cite{he2021beyond} incorporate crowdsourcing with service region design of shared mobility services.

It is crucial to account for the inherent time-dependent uncertainties associated with the adoption of such new services in the service region design and timing decisions with look-ahead policies, particularly for emerging technologies with limited deployment data. The idea is to allow the design to adapt to new conditions while accounting for those possibilities by characterizing the non-stationary uncertainty as a stochastic process. In this context, one approach to consider service region design as a real option is to assume the zones in a region are individual, interacting options featuring multidimensional and non-stationary stochastic origin-destination travel demand. The value of an investment opportunity with the flexibility of deferring individual or subset of zones in the region can be quantified using a real options approach (ROA). 

Therefore, as opposed to a deterministic approach for a static service region design that do not incorporate flexible decisions over multiple time periods, the sequential service region design and timing problem involves adaptive decision making over a decision time horizon. In such case, decisions (driven by the CR policy) are made at each decision epoch (time step) in a rolling time horizon based on the future information available at that epoch and decisions made in previous time steps. This can provide planners with more flexibility in the investment decisions and design of service regions under future uncertainty. The performance within a region can be determined using continuous approximation models (\emph{e.g.}, \citealp{daganzo1986configuration,kim2013integrating}). Moreover, the benefit (investment payoff) arising from such decisions can be defined based on the purpose of the mobility services (\emph{e.g.}, ridership, travel time savings, social welfare, fleet utilization, and carbon emission reduction). For example, considering the MoD services, the investment payoff for a given region can be defined in terms of overall gain in MoD ridership with respect to certain ridership threshold (defined by the operator).
The ridership can be estimated using aggregate network models for urban taxi services (\citealp{salanova,salanova2011review,yang2000macroscopic,wong_taxi}).

\subsection{Real options policy as approximate dynamic programming method for network design and timing}
A real option is a construct used to model flexible decision-making behavior considering volatility of project returns.
%as opposed to the traditional discounted cash flow models, which involve deterministic assumption of returns (\citealp{trigeorgis}). 
%In this sense, real options approach (ROA) provides a way to assess the benefits and opportunities associated to flexible investments (\citealp{chow2010flexible}). 
Based on dynamic programming, by projecting future uncertainty realizations, ROA requires finding the current state of the system in a binomial lattice while conducting a backward induction to determine the optimal strategy at each point in time (or stage). To quantify the value of such flexibility and determine the optimal investment threshold, three numerical methods have primarily been used: finite difference (\citealp{brennan1978finite}), binomial lattice (\citealp{cox1979option}), and Monte Carlo (MC) simulation (\citealp{boyle1977options}). Under multiple and complex uncertain elements, where uncertainties can be modeled as non-stationary stochastic processes, the Monte Carlo simulation method is more suitable compared to the other two methods (\citealp{grant}). However, calculating the expected returns of deferring or exercising an option at each time step, given an expiration time $t_E$, grows increasingly computationally expensive through the backward solution process. 

To address this, \cite{longstaff} present a simple yet cost-effective approach for approximating the option values for timing variables by least-squares Monte Carlo simulation (LSMC). This involves applying least-squares regression to fit expected returns along all simulation paths (realizations). LSMC is used to determine optimal investment threshold (\emph{i.e.,} optimal stopping time) by comparing the expected returns on waiting versus exercising immediately, for each decision time step. LSMC is proven to converge to an approximation of the true option value as the number of paths, time steps, and number of polynomials for regression approach infinity (\citealp{stentoft}). 
This approach is extended by \cite{gamba2003} to a multi-option LSMC approach to handle multiple interacting options. They classify three idealized cases based on the relationship between the options: independent options (value of a portfolio is the sum of each individual option value), compound options (execution of the $h^{th}$ option in a portfolio of $H$ options creates the right to exercise the subsequent $(h+1)^{th}$ option), and mutually exclusive options (exercise of one of the options eliminates the opportunity of execution of the remainder).

Applications of ROA can be found in numerous transportation studies (\citealp{garvin, galera, zhaohighway,couto,huang}). \cite{tibben} provide an overview of ROA applications in logistics and transportation. 
%A review of ROA as a tool for facility location under uncertainty is provided in \cite{snyder}. \cite{garvin, galera, zhaohighway, pichayapan, li2022} apply real option methods to study highway investments under stochastic demand.  ROA has been studied for high-speed railway (HSR) investment evaluation under stochastic HSR demand facing negative or positive shocks (\citealp{couto}), and under minimum revenue guarantee uncertainty (\citealp{huang}). 
Benefits of ROA in the design of Intelligent Transportation Systems is discussed in \cite{de2008real}. 
%\cite{saphores} investigate the application of real options in congestion relief investment focusing on the economic and urban planning aspect. These methods however tend to ignore option interactions with interrelated stochastic flows. 
\cite{chow2011real, chow2011network} develop network-based real option models to consider the timing of investment decisions and project selections under time-dependent uncertainty, resulting in the CR policy (\citealp{chow2016reference}) as a form of value function approximation that includes network design and timing decisions within the broader field of approximate dynamic programming (ADP). \cite{powell2007approximate, powell2021reinforcement} identifies four fundamental classes of policies for decision making using ADP (which in recent years has been generalized within the broader reinforcement learning field): value function approximation (VFA), cost function approximation (CFA), policy function approximation (PFA), and direct look ahead (DLA). In policy approximation solutions, the policy itself is modified. PFA represents a policy which typically does not involve solving an optimization problem and is suited for problem with low-dimensional decisions, while CFA requires solving an optimization problem in a constrained action space and is suitable when a decision is multi-dimensional. On the other hand, VFAs explicitly consider the downstream impact of current decisions over the full decision horizon. VFA represents the value function of a given state in Bellman equations with a functional approximation and works best when the value of the future given a state is easy to approximate. For problems with lack of recognizable structures, DLA is used by directly introducing approximations instead of deriving an explicit function like the other three policies. 

Some recent studies explore ADP based on real options policies in the context of emerging transportation technologies. \cite{chow2016dynamic} uses a hybrid LSMC simulation within a CFA policy (a selective arc inventory routing problem where the reward term is approximated using LSMC) in their study on real-time unmanned aerial vehicle based traffic monitoring.
%under uncertainty formulated as a stochastic arc-inventory routing problem. 
\cite{zheng} investigate ROA for fleet replacement timing decisions for a shipping company under uncertain demand and fuel prices using the CR policy. 
\cite{eng}  provide a qualitative assessment of embedding uncertainty in policy making for autonomous vehicles.
%to develop a range of flexible and adaptive policy options for autonomous vehicles.  
\cite{cardin} apply a decision rule technique using ROA to model flexibility into one-way MoD transportation systems design and operations. The existing literature on the use of ROA in transportation suggests that there is benefit (\emph{i.e.}, added value) in incorporating flexibility to investment decisions under uncertainties. However, for a portfolio of $H$ compounded options, valuating $H!$ investment sequences based on VFA for large-scale cases requires a computationally efficient approach. \cite{chow2016reference} propose a sampling technique to obtain an extreme value distribution of the policy value for the $H!$ sequences without explicit sequence enumeration. The limitation to this approach, however, is that the estimation is not only prone to high variability based on the sample size, but the policy (\emph{i.e.}, the optimal investment strategy) that leads to those values (which can still be used as consistent benchmark reference policy values for evaluating other ADP policies) also remains unknown. 

Therefore, to address the gaps in the literature, this study explores a real options based sequential service region design and timing problem for MoD services. For this approach to be scalable, we propose an efficient neural network based ML method to address the combinatorial complexity arising in the optimal investment strategy and timing decisions for a large number of service zones.

\subsection{Neural networks for sequence understanding} \label{sec:RNN_lit_review}
Understanding a discrete sequence (\emph{i.e.}, with discrete time index), for the purposes of classification or regression, has widespread applications including text understanding (\citealp{liu}), online user behaviour understanding (\citealp{zhou}), and time series forecasting (\citealp{RNN_forecasting_survey}). For text understanding, the underlying sequence stems from the sequences of words in a document, in an online recommendation system, and the user activity history forming the sequence. For applications involving non-trivial sequential patterns, including the ones mentioned above, RNNs have seen significant success (\citealp{deep_learning_book_goodfellow}). RNNs (\citealp{rumelhart, schmidhuber}) are a class of neural networks specifically designed to understand sequences. At a high level, an RNN processes a sequence index by index, and updates its internal state based on the current input and its past state (hence, recurrent in nature). The recurrent unit that processes each value in a sequence can have multiple design choices, including long short term memory (LSTM) (\citealp{schmidhuber}), and gated recurrent unit (GRU) (\citealp{GRU}), both of which try to capture long term memory (GRU is a simplified version of LSTM). With the increasing prominence of data-driven planning, lately, some studies in transportation have begun to seek different variants of RNNs for prediction tasks including forecasting traffic flow (\citealp{bogaerts}), traffic speed (\citealp{ma_speed}), travel time (\citealp{duan_time}), and ride-hailing demand (\citealp{jin}). Applications can also be found in classification of travel modes (\citealp{liu2017end}), travel trajectories (\citealp{liu_spatio}) and traffic crash risk categorization (\citealp{li_crash}). These studies show superior performance compared to traditional statistical learning methods, due to the capabilities of RNNs to process and learn complex sequence patterns.

Motivated by the success of LSTM based models in various transportation studies in the literature, we explore LSTM based RNNs for understanding sequences related to sequential mobility service region design, but our proposed approach can be easily extended to other RNN variants (\emph{e.g.}, GRU, bi-directional LSTMs, RNN with attention (\citealp{deep_learning_book_goodfellow})). LSTMs are a special kind of RNN, capable of learning long-term dependencies. They have been used as a variant of simple recurrent cells in RNN as they have the ability to remove or add information to the cell state, carefully regulated by structures called gates  (\citealp{greff}). The core of LSTM is cell state, which has the ability to regulate information to cells while selectively letting information flow through the door mechanism to achieve this purpose. 
%LSTM consists of three gates: forget gate (decides which information to delete from the cell state), input gate (decides what information to update to the cell state), and output gate (decides the final output of the network). 
A recent development in the natural language processing (NLP) research community, \emph{i.e.}, the introduction of the transformer neural network (\citealp{vaswani}), has led to state-of-the-art results for multiple text understanding tasks. Although they are sequence models by design, they have parameters in the order of hundreds of millions, and require self-supervised learning based pre-training strategies (\citealp{bert}) for effective performance on text data.
Figuring out such pre-training strategies (which do not require labeled data) is non-trivial and is domain dependent.
In this study, we take a step towards leveraging neural sequence models for mobility service region design.
Given the exploratory nature of our work, we focus on RNNs instead of transformers since RNNs have far fewer parameters to learn, and pre-training in not necessarily required to learn non-trivial sequence patterns. 

%%%% METHODOLOGY %%%%
\section{Methodology} \label{sec:methodology}

This section first gives a formal description of the 
sequential mobility service region design and timing problem in \Cref{sec:setup}, \Cref{sec:payoff} describes the investment payoff model used in the study, followed by details of the solution algorithm in \Cref{sec:solution algorithm}. 

\subsection{Problem setup} \label{sec:setup}

Consider an MoD operator planning to implement their services in a new city; the potential service region considered by the operator consists of a (geographically contiguous) set of $\mathcal{H}$ candidate zones ($|\mathcal{H}| = H$). We assume that each service zone $h \in \mathcal{H}$ comprises of multiple sub-zones ($h_{sub}$), such that there are total $\mathcal{H}_{sub}$ sub-zones in the region, % (see \Cref{fig:ROV_components}), 
constituting $\mathcal{H}_{sub}\times\mathcal{H}_{sub}$ origin–destination (OD) pairs. 
Such zone and sub-zone boundaries can be based on statistical areas defined by city planning agencies (e.g., New York City (NYC) has  Public Use Microdata areas (PUMA) that resemble neighborhoods with a minimum of 100,000 population (which can be assumed as zones as per our setup), while taxi zones are relatively smaller statistical areas corresponding to taxi pick-up and drop-off zones (these can be treated as sub-zones aggregated to form PUMA zones); various other zoning criteria can be used by mobility operators and planners where they can group/cluster similar small sub-zones ($\mathcal{H}_{sub}$) into bigger zones ($\mathcal{H}$).
The sequential service region design and timing problem focuses on the long-term multi-period flexible investment planning of new mobility services such as MoD services under non-stationary uncertainty that evolves over time. Let $\mathcal{T}$ be the set of discrete time periods (years) in a finite horizon $t_E$ ($\mathcal{T}$ = $\{t_1,t_2,\dots,t_E\}$), such that $t_n$ is the $n^{th}$ time step in $\mathcal{T}$. We assume a rolling time horizon of $E$ years, such that, for each time step or decision epoch $t_d$ in the horizon, a roll period of $\mathcal{T}$ (future) time periods starting from $t_d$ is considered. The underlying assets in our setup are the $\mathcal{H}$ interacting service zones. The objective is to determine the optimal real options based investment strategy for a candidate set of zones ($\mathcal{H}_{cand}$ $\subseteq \mathcal{H}$) at each decision epoch (\emph{i.e.}, invest immediately or defer, where the defer decision at the final time period of a horizon is equivalent to rejecting the decision) based on the (future) information available at the epoch and the zone investment decisions from previous time step (as illustrated in \Cref{fig:illustrative_example}) for an adaptive service region design. Therefore, at each decision epoch $t_d$, the aim is to determine the investment/deferral strategy  $a_h \colon \{1,0\}$ for each $h \in$ $\mathcal{H}_{cand}$. The following sections describe the proposed method for the service region design and timing of $\mathcal{H}$ service zones for a single decision epoch in the rolling horizon. For simplicity, we denote the decision epoch as $t_0$ with $\mathcal{T}$ discrete time periods ($\mathcal{T}$ = $\{t_1,t_2,\dots,t_E\}$).

%The underlying assets (projects) of the proposed multi-option model are the interacting zones ($H$) of the service region. At a decision epoch, The projects can be invested or deferred any year from the initial year $t_0$ to the maturity $t_E$. 

Although various sources of uncertainty exists for new transportation systems, for the long-term planning of such services, OD demand generally exhibit greatest uncertainty (\citealp{chow2010flexible}). Hence, the main source of uncertainty considered in our study is the overall OD demand which includes travel within and between specific service zones to be served by the system. We assume that this accounts for the uncertainty in the mobility service adoption along with external market factors together influencing the evolution of service demand over time. 

Suppose that $Q_{ijt} \in \mathbb{R}^{|\mathcal{H}_{sub}|\times|\mathcal{H}_{sub}|}$ be the underlying OD demand for the MoD service at time $t_n \in \mathcal{T}$ (for a specific OD pair $ij \in \mathcal{H}_{sub}\times\mathcal{H}_{sub}$); this can be characterized by a stochastic process, such as a Geometric Brownian motion (GBM). GBM is a continuous-time stochastic
process in which the logarithm of the variable follows a Brownian motion (\citealp{ross}), and has been used in the literature to model travel demand and vehicular demand (\citealp{zhaohighway, li2015transit, chow2011network, saphores, gao}). Based on this, we model the OD demand in our study as a GBM process; this is assumed to be independent between zones (\citealp{chow2011network}) although correlated multivariate processes can also be considered with appropriate data. Under this assumption the service demand at each OD pair $ij$ satisfies \Cref{eq:weiner}.

\begin{align}
    \dfrac{dQ_{ijt}}{Q_{ijt}} = \mu Q_{ijt}dt + \sigma Q_{ijt}dW_t \label{eq:weiner}
\end{align}

where $dt$ an infinitesimal time increment, $\mu$ is the drift, $\sigma$ is the volatility rate of the service demand, and $dW_t \sim N(0,dt)$ is a standard Wiener process.\footnote{A Wiener process is a real valued continuous-time stochastic process with three important characteristics. First, it is a Markov process; second, it has independent increments; and third, changes in the process over any finite interval of time are normally distributed.} However, it is less likely that the demand for all OD pairs will have heterogeneous volatility. Hence, we assume zone-specific volatility ($\sigma_H: \{\sigma_1, \sigma_2,...,\sigma_H\}$ for $H$ zones), such that all OD pairs originating from zone $h$ (\emph{i.e.}, $h_{sub}\times\mathcal{H}_{sub}$) have $\sigma_h$ volatility. 

Therefore, for a decision epoch ($t_0$), the objective is to determine the optimal investment strategy for $H$ service zones subject to the compound options under demand uncertainty and heterogeneous volatility, where investment in each zone can be made immediately or deferred (until $t_E$) to maximize the value of the portfolio. As discussed earlier (in \Cref{sec:introduction}), for a portfolio of H compounded options, $H!$ possible investment sequences can be constructed. The benefit or performance of a selected set of zones in a region is evaluated using investment payoff. \Cref{sec:payoff} below describes the payoff calculation used in our study.

%%%%%%%%%%%%%%%%%%%%%%%%%%%%%%%%%%%%%%%%%%%%%%%%%%%%%%%%%%%%%%%%%%%%
\subsection{Investment payoff calculation} \label{sec:payoff}
%As discussed earlier, we assume the OD demand across $\mathcal{H}_{sub}$ subzones (in a region with $H$ zones) is the stochastic variable that evolves over time. 
As discussed in \Cref{sec:related_work}, the investment payoff in our setup is defined based on MoD ridership. Let $Q_{ij,t_n}$ represent the potential (peak hour) demand for MoD services between $\mathcal{H}_{sub}$ sub-zones ($i \in \mathcal{H}_{sub}, j \in \mathcal{H}_{sub}$ in a region with $\mathcal{H}$ zones) at a certain time $t_n \in \mathcal{T}$. The MoD ridership across different service zones ($\mathcal{H}$) in a region can be estimated based on the potential service demand and other region characteristics. We adopt the aggregate network model for urban taxi services by \cite{wong_taxi} to estimate OD ridership for the MoD services in our setup. As shown in \Cref{eq:od_ridership}, we use an exponential function (\citealp{wong_taxi}) for expressing hourly MoD ridership $\lambda_{ij}$ for an OD pair $ij$.
\begin{align}
    \lambda_{ij} = Q_{ij,t_n} \exp^{-\gamma(c_{ij}+\alpha_{IV}.VoT.TIV_{ij}+\alpha_W.TW)} \label{eq:od_ridership}
\end{align}

where $VoT$ is the value of time of MoD users in the service region ($euro/min$), $TIV_{ij}$ is the in-vehicle travel time ($min$) between OD $ij$, $c_{ij}$ is the trip price (in $euros$) for OD $ij$, $TW$ is the expected wait time ($min$), $\gamma$ is the congestion scaling parameter ([0,1]); $\alpha_{IV}$ and $\alpha_W$ are customer perception factors for in-vehicle time and wait time respectively. 
For simplicity, we assume a taxi dispatching market where the number of vehicles is optimum, and use the following formulation in \Cref{eq:wait_time} for estimating the wait time (\citealp{salanova}).
\begin{align}
    TW =  0.8\lambda_u^{1/3}v^{-2/3} \label{eq:wait_time}
\end{align}

where $v$ is the average vehicle speed ($km/hr$) and $\lambda_u$ is the hourly MoD ridership in a region given by \Cref{eq:reg_ridership}.
\begin{align}
    \lambda_u = \sum_i \sum_j \lambda_{ij}, \;\; i \in \mathcal{H}_{sub}, j \in \mathcal{H}_{sub} \label{eq:reg_ridership}
\end{align}

Therefore, based on the formulations above, for a given set of $\mathcal{Z}$ zones ($\mathcal{Z} \subseteq \mathcal{H}$) with $\mathcal{Z}_{sub}$ sub-zones ($\mathcal{Z}_{sub} \subseteq \mathcal{H}_{sub}$) constituting $\mathcal{Z}_{sub}\times\mathcal{Z}_{sub}$ OD pairs, we calculate the cumulative (peak hour) MoD ridership using \Cref{alg:ridership_algo}.
\begin{algorithm}[H] 
\begin{algorithmic}[1]
%\begin{ALC@g}
\STATE Initialize $\lambda_{0,ij} = Q_{ij,t_n}$; $i, j \in \mathcal{Z}_{sub}$\\
\STATE Calculate $\lambda_{0,u} = f(\lambda_{0,ij})$;  $i, j \in \mathcal{Z}_{sub}$ using \Cref{eq:reg_ridership} 
\STATE Calculate $TW_0$ = $f(\lambda_{0,u}, v)$ using \Cref{eq:wait_time}\\
\STATE Initialize $gap_{wait}$ = 1000 ($min$), $itr$ = 0, $tol$ = 0.001 ($min$)\\
\STATE \While{$gap_{wait} \geq tol$}
{
    \Indp \STATE Update $itr = itr + 1$ 
          \STATE Calculate $\lambda_{itr,ij} = f(\lambda_{itr-1,ij},TW_{itr-1}, c_{ij}, VoT, TIV_{ij}, \gamma, \alpha_{IV}, \alpha_{W})$; $i, j \in \mathcal{Z}_{sub}$ using \Cref{eq:od_ridership}
          \STATE Calculate $\lambda_{itr,u} = f(\lambda_{itr,ij})$;  $i, j \in \mathcal{Z}_{sub}$ using \Cref{eq:reg_ridership}
          \STATE Calculate $TW_{itr} = f(\lambda_{itr,u}, v)$ using \Cref{eq:wait_time}
         \STATE Update $gap_{wait} = TW_{it}-TW_{itr-1}$
}
\STATE Cumulative ridership for $Z$ zones = $\lambda_{itr,u}$; OD ridership for $\mathcal{Z}_{sub}$ sub-zones in $Z$ zones = $\lambda_{itr,ij}$
\end{algorithmic}
\caption{MoD ridership calculation for a subset of $Z$ service zones in a region with $\mathcal{Z}_{sub}$ sub-zones} \label{alg:ridership_algo}
\end{algorithm}%}%

%The optimal investment strategy for $H$ service zones with $\mathcal{H}_{sub}$ sub-zones subject to the compound options is solved using Algorithm 2. 
%As shown in Algorithm 2 (steps 6-8), for a service zone $z_h$ in a sequence $s:\{z_1,z_2,...,z_H\}$, 
Consider an investment sequence $s:\{z_1,z_2,...,z_H\}$ ($|s| = H$).  For a service zone $z_h \in s$, the cumulative ridership from addition of $z_h$ in a region with $z_1,z_2..,z_{h-1}$ zones is obtained using \Cref{alg:ridership_algo} (let's say $X^{cum}_{z_h}$). Therefore, the service zone ridership for $z_h$ (\emph{i.e.}, $X_{z_h}$) can be calculated as $X^{cum}_{z_h} - X^{cum}_{z_{h-1}}$. 
%Now, based on the OD ridership for the sub-zones in $\{z_1,z_2,..z_h\}$ (obtained from \Cref{alg:ridership_algo} ), we calculate the 
The investment payoff for zone $z_h$ (\emph{i.e.,} $\pi_{z_h}$) is calculated as per \Cref{eq:zone_payoff}. 
\begin{align}
    \pi_{z_h} =  X_{z_h} - (C_{wz} + 2(h-1)C_{iz}) \label{eq:zone_payoff}
\end{align}

Essentially, we assume the payoff (benefit) from an investment in a zone $z_h$ is the ridership gained from including zone $z_h$ in the service region (\emph{i.e.}, $X_{z_h}$) minus a ridership threshold (similar to a strike price in stock options (\citealp{longstaff})). The ridership threshold represents the cost component in \Cref{eq:zone_payoff}. This includes a pre-defined within-zone cost $C_{wz}$ for within-zone riders and interzone cost $C_{iz}$ for interzone riders. Based on the order or position of zone $z_h$ in a sequence, the total interzone cost becomes $2(h-1)C_{iz}$, since it opens up $2(h-1)$ new connections (\emph{i.e.}, interzone links) with the preceding $h-1$ zones in the sequence. For example, consider a sequence $\{z_1,z_2,z_3\}$ for a region with three zones. Using \Cref{eq:zone_payoff}, starting with $z_1$ we only have $C_{wz}$ as the (within-zone) cost, then, adding $z_2$ opens up $2$ interzone links (\emph{i.e.}, $z_1-z_2, z_2-z_1$), therefore cost becomes $C_{wz}+2C_{iz}$. Next, by adding $z_3$, we get 4 new interzone links (\emph{i.e.}, $z_3-z_2,z_3-z_1,z_2-z_3,z_1-z_3$), so the cost is $C_{wz}+4C_{iz}$. This approach simplifies the benefit/cost evaluation to use units of ridership as a proxy of revenue and cost without having to assume monetary conversions.

\subsection{Solution algorithm for service region design and timing decisions} \label{sec:solution algorithm}
%% SETUP %%
For a portfolio of $H$ interacting service zones with $H!$ possible investment sequences (where sequences follow the features of the compound options described in \cite{gamba2003}), let $\mathcal{S}$ denote a set constituting $H!$ sequences and $L$ represents the total number of such sequences. To determine the optimal investment strategy subject to the compound options, we adopt the LSMC method (based on \citealp{gamba2003, chow2011network}) for estimating the policy value of each investment sequence $s \in \mathcal{S}$. Specifically, this method integrates MC simulations and combines a forward-looking model for incorporating uncertainties along with a backward recursion (\emph{i.e.,} least-squares regression) for accurately estimating the value of an option. For a given set of $\mathcal{H}$ zones and set of zone volatility $\sigma_H$, each of the $\mathcal{H}_{sub}\times\mathcal{H}_{sub}$ OD pair service demand is simulated by generating a set of MC realization paths (let this be denoted as $\mathcal{P}$) by means of GBM for each $t_n \in \mathcal{T}$. The continuation value at each $t_n$ is then estimated, which is essentially an estimate of the investment value of the next time step ($t_{n+1}$), given a realization $p \in \mathcal{P}$. This denotes the value of continuing to wait for the realization of future OD demand at each time along the $p^{th}$ path. The optimal stopping time along each MC path is found by comparing the continuation value to the value of investing immediately. In this setup, one assumption is that the budget or funding remains available over the time horizon and can be used for the investment at a later time. 

We adapt the CR policy (\citealp{chow2011network,chow2016reference,chow2018informed}) as a VFA method in approximate dynamic programming (\citealp{powell2007approximate}) to solve the lower bound
MoD service region design and timing problem with stochastic variables. This is used to estimate the option value for each of the enumerated sequences in $\mathcal{S}$. The goal is to search for the sequence that gives the highest option value (\citealp{chow2011network}) which forms the optimal investment strategy (\emph{i.e.,} invest now or defer later for each zone $h \in \mathcal{H}$). Details of the CR policy is included in \Cref{sec:CR policy}. The challenge, however, is that there is increasingly high computational cost in determining the policy value for all $H!$ sequences for a large number of zones. To address this computational complexity, we present an ML based CR-RNN policy to accelerate the above calculations for efficiently determining the optimal investment strategy for the service region design. \Cref{sec:CR RNN policy} includes the proposed CR-RNN policy (solution algorithm in \Cref{alg:cr_rnn_algo}) and details of the RNN component in the proposed method. \\

%%%%%%%%%%%%%%%%%%%%%%%%%%%% CR policy %%%%%%%%%%%%%%%%%%%%%%%%%%%%%%%%%%%%%%%%%%%%%%
\subsubsection{CR policy for the service region design and timing decisions} \label{sec:CR policy}
\;\\

\begin{figure}[!htb]
\begin{center}
\includegraphics[scale=.45]{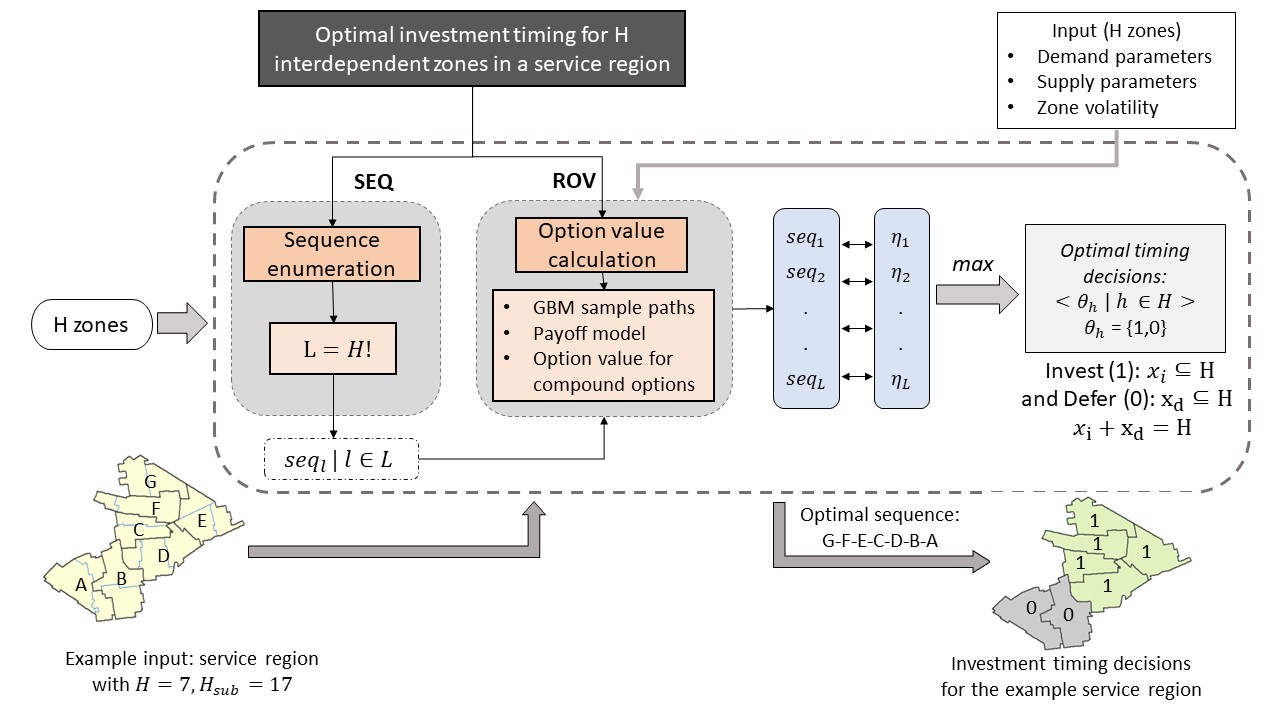}
\caption{CR-RNN policy model for determining which mobility service region zones to serve now and which to defer (zones and sub-zones illustrated in the figure are PUMA and taxi zones in NYC respectively).} 
\label{fig:ROV_components} 
\end{center}
\end{figure}

\Cref{fig:ROV_components} provides an illustration of the steps involved in the sequence policy value estimation using multi-step compound option valuation based on the CR policy. As shown in the figure, this can be categorized into two main components:
\begin{itemize}
    \item $SEQ$: This includes enumeration of possible $H!$ permutations (\emph{i.e.}, ordered sequences in set $\mathcal{S}$) for $H$ zones ($L = |\mathcal{S}|$)
    
    \item $ROV$: This involves real option valuation of each sequence in $\mathcal{S}$. The ROV component majorly contributes to the overall computation time in the RO model, and hence, is the focus of our study. 
\end{itemize}

Consider an investment sequence $s \in \mathcal{S}$ where $s = \{z_1, z_2,...,z_H\}$ ($|s| = H$). For multi-step compound options (\citealp{gamba2003}), the value of the subsequent options is incorporated when valuing the deferral option for any project $z_h \in s$. The result is a Bellman equation for determining the value of the option to invest in a service region design as a function of stochastic OD demand:
\begin{align}
    F_{z_h}(t_n, X_{z_h,t_n}) = \text{max} \{\pi_{z_h}(t_n, X_{z_h,t_n}) + F_{z_{(h+1)}}(t_n, X_{z_h,t_n}), (1+\rho)^{-(t_{n+1}-t_n)} \mathbb{E}_{t_n} [F_{z_h}(t_{n+1}, X_{z_h,t_{n+1}})]\}, \label{eq:bellman}
\end{align}

where $X_{z_h,t_n}$ is state variable of the project. In our setup, $X_{z_h,t_n}$ represents the estimated MoD ridership in zone $z_h$ at $t_n$, which is a function of the stochastic OD demand (in the region) at time $t_n$ (obtained using \Cref{alg:ridership_algo}),
%as shown in \Cref{eq:rider} 
%	\begin{align}
%	    X_{z_h,t_n} = f(Q_{ijt_n}) \; \forall ij \in \mathcal{H}_{sub}\times\mathcal{H}_{sub} \label{eq:rider}
%	\end{align}
	
$\rho$ is the risk-free discount rate\footnote{For the purposes of this research, we assume future costs are risk-free},\\
\indent $\pi_{z_h}(t_n, X_{z_h,t_n})$ is the net present worth of an immediate investment of a project $z_h$ at time step $t_n$. Here, the total investment payoff of a zone $z_h$ equals the payoff from immediate investment of $z_h$ (obtained using \Cref{eq:zone_payoff} with $X_{z_h}$) plus the opportunity of investing in $z_{(h+1)}$ project (\emph{i.e.}, $F_{z_{(h+1)}}(t_n, X_{z_h,t_n}$)),\\
\indent $\mathbb{E}_{t_n} [F_{z_h}(t_{n+1}, X_{z_h,t_{n+1}})]$ is the expectation function of all future investment benefits from the contingent claim optimally exercised at a certain time. Hence, it is the option value for zone $z_h$ at time step $t_{n+1}$ where the expectation is taken conditional on the information available at time $t_n$. Therefore, the second term in \Cref{eq:bellman} represents the continuation value (let's denote this by $\phi_{z_h}(t_n,X_{z_h,t_n})$) and is estimated by regressing the discounted future investment values on a linear combination of a group of basis functions of the current state variables ($X_{z_h,t_n}$).

We use hermite polynomials to define the basis functions (\citealp{,gamba2003, longstaff}) such that $L_j$ is the orthonormal basis of the $j^{th}$ state variable, and $J$ is the number of basis functions assumed. \Cref{eq:cont_val,eq:payoff_contval,eq:regression_approx} show the continuation value calculation of a service zone $z_h \in s$ at time step $t_n$ for a sample path $p$. 

\begin{align}
    \phi_{z_h}(t_n,X_{z_h,t_n}(p)) &= \mathbb{E}_{t_n}^*\bigg[\sum_{i=n+1}^E(1+\rho)^{-(t_i-t_n)} \sum_{r=h}^{H} \pi_{z_r}(t_n,t_i,\tau,p)\bigg] \label{eq:cont_val}\\
    \text{with} \; \;	\pi_{z_h}(t_n,t_i,\tau,p) &= 
            \begin{cases}
                    \pi_{z_h}(t_i,X_{z_h,t_i} (p),& \text{if } \; t_i= \tau_{z_h}(p)\\
                              0,    & \text{otherwise}
            \end{cases} \label{eq:payoff_contval}\\
    \implies \phi_{z_h}(t_n,X_{z_h,t_n}(p)) &\approx \phi_{z_h}^*(t_n,X_{z_h,t_n}(p)) = \sum_{j=1}^J\beta^*_j(t_n)L_j(X_{z_h,t_n}(p)) \label{eq:regression_approx}
\end{align}

\indent where $\beta^*_j$  are the optimal coefficients for the $J$ basis functions obtained using least-squares estimation (\citealp{longstaff}) using all paths (\citealp{gamba2003}).

Let $\tau_{z_h}(p)$ be the optimal stopping time for a zone $z_h$ in the $p^{th}$ path, and $\theta_{z_h,t_n}$ be the decision to invest or exercise immediately in zone $z_h$ at time $t_n$ ($\theta_{z_h,t_n}$ = \{1: invest, 0: defer for later\}). Along each path, if the payoff of $z_h$ exceeds the continuation value at $t_n$, the investment is exercised and the optimal stopping time is updated to $t_n$; this is done recursively from maturity $t_E$ to $t_1$ (see \Cref{alg:cr_rnn_algo} steps 11-17). 
As noted by \cite{chow2011network}, in the multi-option cases, a backward option recursion is added to the backward time recursion. Therefore, using the optimal stopping times of each MC path ($p \in \mathcal{P}$), which includes the earliest investment timing of each zone $z_h \in s$, the option value at $t_0$ is determined (as shown in \Cref{alg:cr_rnn_algo} steps 18-21). 
Finally, the investment sequence that offers the highest option value for the initial project ($\eta_s$ denotes policy value of a sequence $s$) is selected. %(Algorithm 2 step 20). 
The corresponding policy can be used as a reference for the optimal investment decision (\emph{i.e.,} invest now or defer for each zone $h \in \mathcal{H}$). This considers managerial flexibility and future uncertainties in deciding the optimal investment strategy. Note that the best sequence is not necessarily the final investment strategy but serves as the lower bound on the true solution.

As noted by \cite{chow2011network}, some of the sub-sequences in $H!$ are repeated and the cumulative benefits (\emph{i.e.,} ridership and payoff) of the option $z_h$ for a sequence $s_1$ at time
$t_n$ can be re-used for an option in another sequence $s_2$. Therefore, the effective computational cost of evaluating $H!$ sequences is on the order of $O\bigg(\mathlarger{\mathlarger{\sum^{H}_{h=1}}} \dfrac{H!}{H-h!}\times|\mathcal{P}|\times|\mathcal{T}|\times \text{payoff(H)}\bigg)$.
%Note that, the objective of the payoff model in our study is to get an estimate of MoD ridership and payoffs; the design or modeling of MoD services is not the focus of this paper. 
The above framework allows for various other stochastic variables that can be considered by planners based on their primary concerns. Our study assumes exogenous demand and zone volatility, but in practice their interaction with the investment decisions need to be also considered in the modeling process. Moreover, in real project implementation, models for optimizing supply-demand parameters in a multi-modal system based on different objectives and operation policies (\citealp{wei2020modeling,djavadian,chow2021urban, liu_mod, pantelidis2020many}) can be used to get better estimates for the investment payoffs. The complexity of such payoff models in the ROV calculations makes it even more computationally expensive for large-scale decision making. The section below describes the proposed CR-RNN policy which includes a computationally efficient approach to obtaining the optimal investment strategy.
\bigskip
%%%%%%%%%%%%%%%%%%%%%%%%%%%% CR policy %%%%%%%%%%%%%%%%%%%%%%%%%%%%%%%%%%%%%%%%%%%%%%
\subsubsection{CR-RNN policy for efficient service region design and timing decisions} \label{sec:CR RNN policy}
\;\\

The objective of the CR-RNN policy is to efficiently obtain the investment sequence out of $H!$ sequences that offers the highest initial project value, which forms the optimal investment strategy \emph{i.e.,} invest now or defer later for each candidate service zone $h \in H$. \Cref{alg:cr_rnn_algo} includes the details of the solution algorithm for the CR-RNN policy. 

\begin{figure}[!htb]
\begin{center}
\includegraphics[scale=.50]{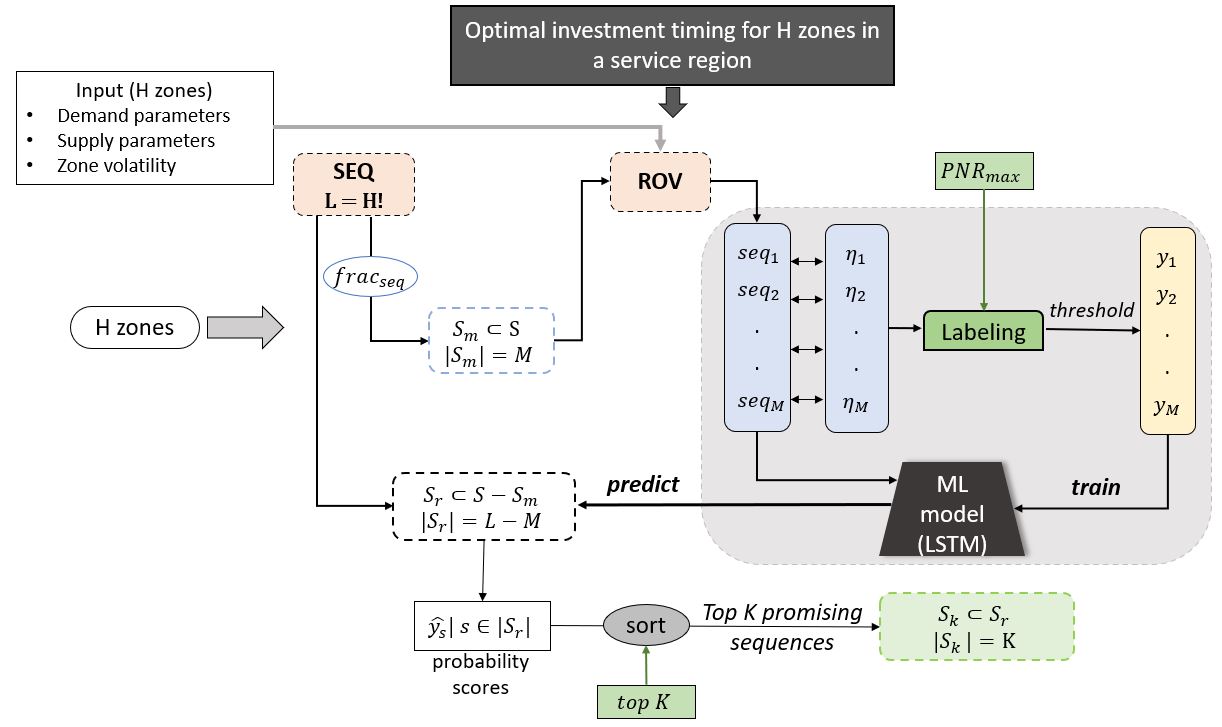}
\caption{Using the RNN based sequence classification in the CR-RNN policy for efficient ROV calculations} 
\label{fig:lstm_plus_rov} 
\end{center}
\end{figure}

Given a set of $\mathcal{H}$ zones with $H!$ sequences in set $\mathcal{S}$ ($|\mathcal{S}| = L = H!$), 
%we randomly sample a small fraction of sequences ($frac_{seq}$) from $S$ to train the RNN model. Let $S_{m}$ denote the selected subset of sequences such that $|S_{m}|$ = $M$ ($M = frac_{seq}\times L$). 
the idea of the proposed approach (as illustrated in \Cref{fig:lstm_plus_rov}) is to randomly sample a small fraction of sequences ($frac_{seq}$) from the population set $\mathcal{S}$. Let $\mathcal{S}_{m}$ denote the selected subset of sequences such that $|\mathcal{S}_{m}|$ = $M$ ($M = frac_{seq}\times L$). This is used to train an RNN model that learns patterns across different sequences to be able to identify a small set of promising sequences (\emph{i.e.,} sequences with high policy values). In particular, we frame this problem as a sequence classification problem, where we use a labeling strategy (details in Section~\ref{sec:labeling}) to label $M$ sequences (as 1s and 0s) depending on their policy values relative to a threshold defined from the reference policy in \cite{chow2016reference} (\Cref{alg:cr_rnn_algo} step 23). This is used to train the RNN model that effectively learns which type of zone orderings or sub-sequence patterns result in high policy values; identifying such patterns are non-trivial especially for a large set of zones. 

The trained model is then applied on remaining $\mathcal{S}_r$ sequences ($|\mathcal{S}_r| = L-M$) to get the probability estimate of how likely a sequence in $\mathcal{S}_r$ is a promising sequence (\Cref{alg:cr_rnn_algo} steps 24-25). The output of the model \emph{i.e.}, the probability scores ($\hat{y}_s$, $s \in L-M$) are used as ranking scores to retrieve top $K$ sequences (as illustrated in \Cref{fig:lstm_plus_rov}). In particular, the $\mathcal{S}_r$ sequences are sorted in descending order of their ranking scores, and top $K$ sequences $\mathcal{S}_k$ (\emph{e.g.}, top $30$ out of thousands of sequences) are selected; ROV calculations are performed only on these $\mathcal{S}_k$ sequences to obtain their policy values (\Cref{alg:cr_rnn_algo} steps 26-27). Accordingly, the optimal investment sequence for $H$ zones (\emph{i.e.}, one that offers the highest policy value) can be obtained using $\mathcal{S}_{m}$ and $\mathcal{S}_{k}$ sequences only (step 28 \Cref{alg:cr_rnn_algo}). The corresponding policy ($\theta_{z_h,t_0}$ values) is used as a reference for the optimal investment decision for the $H$ zones. Therefore, if the total computational cost for $L$ sequences without the RNN component in the proposed method (\emph{i.e.}, via CR policy) is $C_L$, by using the CR-RNN policy, we achieve significant cost reduction by eliminating the ROV calculations for $L-(M+K)$ sequences ($M+K << L$); this accelerates the computations involved in determining the optimal investment strategy for the service region design.
%This saves significant computation time, which would otherwise be required in the RO valuation of all $L-M$ sequences (without the proposed ML method), compared to computing only top $K$ sequences efficiently selected using the proposed CR-RNN policy ($M+K << L$). 

\;\\
%{\SetAlgoNoLine%
\begin{algorithm}[H] 
\begin{algorithmic}[1]
%\begin{ALC@g}
\STATE Input: $H$ service zones, $\mathcal{H}_{sub}$, $\sigma_H$, $C_{wz}$, $C_{iz}$, $D^\sim_{ij}, c_{ij}, VoT, TIV_{ij}, v, \gamma, \alpha_{IV}, \alpha_{W}$, $frac_{seq}$, top $K$\\
\STATE Generate GBM paths for $\mathcal{H}_{sub}\times\mathcal{H}_{sub}$ OD demand for each path $p \in \mathcal{P}$ = $\{1,2,3,\dots,P\}$ for each time step $t_n \in \mathcal{T}$ (as per \Cref{eq:weiner})\\
\STATE Initialize a dictionary $seq_{\eta}$ to store sequence and their policy values \\
\STATE Get all possible ordered set of sequences $S$ = H! sequences\\
\STATE Randomly sample $\mathcal{S}_m$ sequences from $S$ (using $frac_{seq}$)\\
\STATE \For{sequence $s \in S_m$, where $s = \{z_1,z_2,\dots,z_H\}$}
   {
    \Indp \STATE \For{zone $h = z_1:z_H$}{
        \STATE Estimate cumulative ridership using \Cref{alg:ridership_algo} (with $Z$ = $\{z_1,\dots,z_h\}$):  $X_{z_h,t_n}^C (p)$, $\forall \; t_n \in \mathcal{T}, p \in \mathcal{P}$\\ 
        \STATE Get ridership of the $h^{th}$ zone: $X_{z_h,t_n}(p)= X_{h,t_n}^C (p)-X_{z_{h-1},t_n}^C (p)$, $\forall \; t_n \in \mathcal{T}, p \in \mathcal{P}$
        \STATE 	Get payoff value of the $h^{th}$ zone: $\pi_{z_h}(t_n, X_{z_h,t_n} (p))$, $\forall \; t_n \in \mathcal{T}, p \in \mathcal{P}$ using \Cref{eq:zone_payoff}\\
        }
        \STATE Initiate $h = H, \; z_h = z_H$\\
        \STATE \For {$t_n=t_E:t_1$}{
            \STATE \While{$h > 0$}
            {
            	\STATE Calculate $\phi_{z_h}(t_n, X_{z_h,t_n} (p))$ $\forall$\; $p \in \mathcal{P}$ (Least-squares estimation as in \Cref{eq:cont_val,eq:payoff_contval,eq:regression_approx}) 
            	\STATE \If{$\pi_{z_h}(t_n, X_{z_h,t_n} (p)) + F_{z_{(h+1)}} (t_n,X_{z_h,t_n} (p)) \geq \phi_{z_h}(t_n,X_{z_h,t_n}(p))$}
            	{
            		(a) $\theta_{z_h,t_n} = 1; \tau_{z_h}(p)=t_n$\\
            		(b) $F_{z_h}(t_n,X_{z_h,t_n} (p))= \pi_{z_h} (t_n,X_{z_h,t_n} (p)) +F_{z_{(h+1)}} (t_n,X_{z_h,t_n} (p))$
            	}
            	\STATE \Else 
            	{
            	    (a)	$\theta_{z_m,t_n} = 0, \; \forall \;  m \in \{h,\dots,H\}$\\
            	    (b) $F_{z_m}(t_n,X_{z_m,t_n}(p))=(1+\rho)^{-(t_{n+1}-t_n)} (F_{z_m}(t_{n+1},X_{z_m,t_{n+1}} (p))),  \; \forall \; m \in \{h,\dots,H\}$
            	}
            \STATE h = h - 1
            }
        }
        \STATE $F_{z_h} (t_0,X_{z_h,t_0})=  \dfrac{1}{P} \sum_{w=1}^P (1+\rho)^{-\tau_{z_h}(w)} F_{z_h} (\tau_{z_h}(w),X_{h,\tau_{z_h}(w)}(w)), \; \; \forall \; z_h \in s$
        \STATE \For{$h = H:1$}
            {
            \STATE \If{$\pi_{z_h}(t_0,X_{z_h,t_0 })+ F_{z_{(h+1)}}(t_0,X_{z_h,t_0}) \geq F_{z_h}(t_0,X_{z_h,t_0})$}
                {
                Invest $\rightarrow \theta_{z_h,t_0} = 1; \;
                F_{z_h}(t_0,X_{z_h,t_0}) = \pi_{z_h}(t_0,X_{z_h,t_0 })+F_{z_{(h+1)}}(t_0,X_{h,t_0})$
                }
            \STATE \Else 
             {
                Defer $\rightarrow \theta_{z_m,t_0} = 0, \; \forall \;  m \in \{h,\dots,H\}$
                }
         }
        \STATE $\eta_s = F_{s,h=1}(t_0, X_{z_h,t_0}); \;$ $\text{Add s:}\eta_s \; \text{to} \; seq_{\eta}$
        }
\STATE Label $\mathcal{S}_m$ sequences in $seq_{\eta}$ as per Section~\ref{sec:lstm arch} to Section~\ref{sec:labeling} 
\STATE Train the RNN classifier using $\mathcal{S}_m$  labeled  sequences in step 23
\STATE Apply the trained RNN model to $S_r$ sequences ($S_r = S - S_m$) to get the probability scores $\hat{y}_s \;\forall s \in S_r$
\STATE Select top $K$ sequences ($S_k$) using the scores from step 25
\STATE Apply steps 6 - 22 on $S_k$ sequences 
\STATE $s_{opt}$ = $argmax_s\{seq_{\eta}\}$
\end{algorithmic}

\caption{CR-RNN policy algorithm for optimal investment timing in service region design with $H$ service zones} \label{alg:cr_rnn_algo}
\end{algorithm}%}%

For the CR policy, steps 6-22 in \Cref{alg:cr_rnn_algo} are applied on all $H!$ sequences in the population set $\mathcal{S}$, and accordingly the sequence that offers the highest initial option value is selected (\Cref{alg:cr_rnn_algo} step 28). The sections below describe the training and evaluation process of the proposed RNN model for sequence classification.\\

\begin{figure}[!htb]
\begin{center}
\includegraphics[scale=.53]{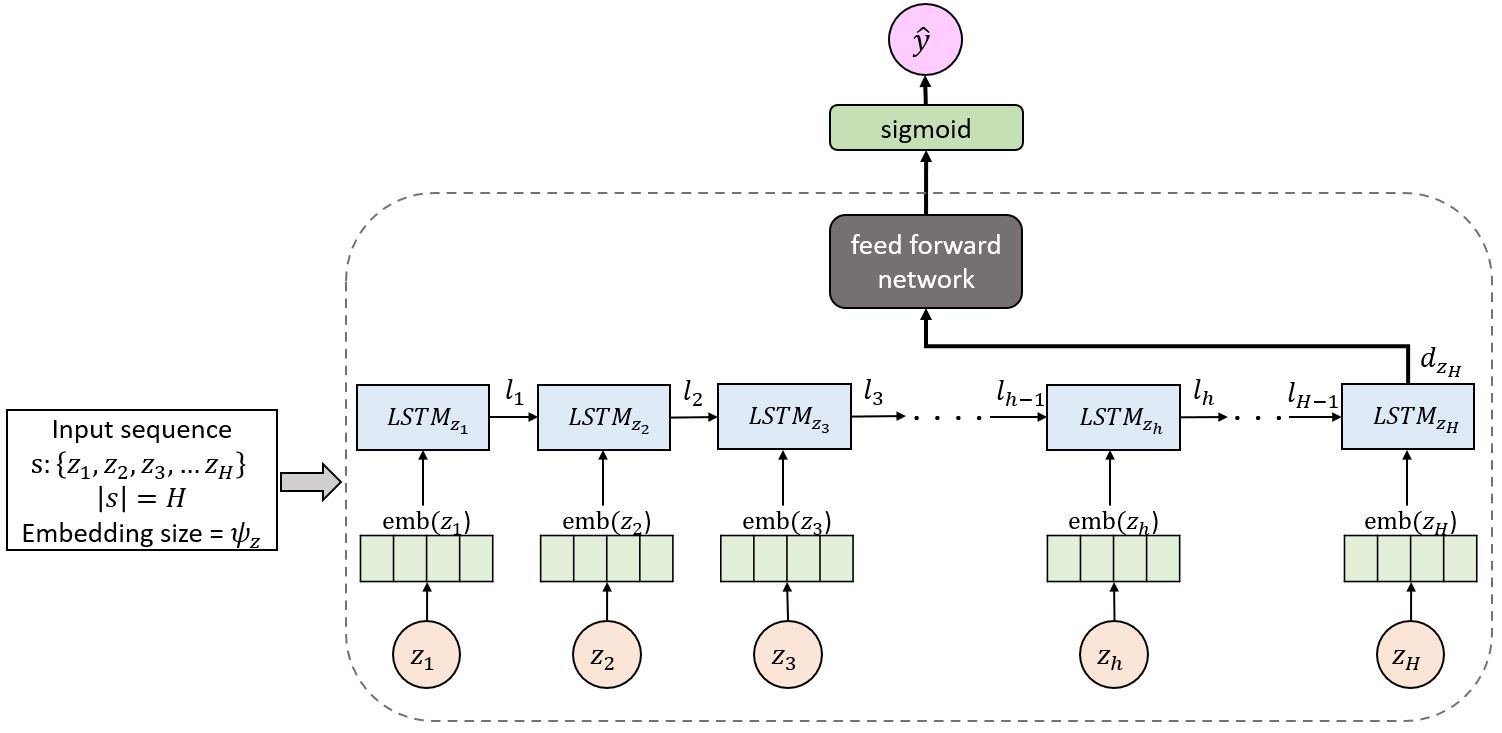}
\caption{LSTM based RNN architecture for investment sequence classification.} 
\label{fig:lstm_architecture} 
\end{center}
\end{figure}

\paragraph{RNN based sequence classification} \label{sec:lstm arch}
\Cref{fig:lstm_architecture} shows the LSTM based RNN model architecture for classifying a candidate sequence $s$ as a promising sequence.
The model has the  following components listed in the bottom-to-top order.
\begin{itemize}
    \item Input embedding layer: this layer takes as input the zone ID ($z_h$), and looks up the embedding (vector representation) for $z_h$ denoted by $emb(z_h) \in \mathbb{R}^{\psi_z}$, where $\psi_z$ is the embedding size (an input to the model). The embeddings are learnt as part of model training in our setup.
    \item LSTM layer: the LSTM layer has $H$ LSTM units. The LSTM unit for index $h$, takes as inputs the embedding for $z_h$, and signals from the LSTM unit for index $h-1$ as shown in \Cref{fig:lstm_architecture}. In other words, the hidden state of $LSTM_{z_h}$ depends on both the current input $emb(z_h)$ and the previous unit $LSTM_{z_{h-1}}$'s state. We follow the standard update rules for LSTM state updates (\citealp{deep_learning_book_goodfellow}) and the details are described in Appendix~\ref{appendix:lstm_appendix}.
    The output of the last LSTM unit is denoted by $l_{z_H}$ and it has two components: the hidden state ($d_{z_H}$) and the cell state ($c_{z_H}$); details around the difference in the cell state and hidden state are in Appendix~\ref{appendix:lstm_appendix}. We just use the hidden state for the subsequent feed forward network (described below) as the hidden state is expected to represent the entire input sequence.
    
    \item Feed forward layer: the hidden state $d_{z_H}$ of the last LSTM unit is fed to a feed forward neural network (single layer) as shown in Figure~\ref{fig:lstm_architecture}. Specifically,
    \begin{align}
        l_{FF} = W_{FF} d_{z_H}  + b_{FF} \label{eq:ffn},
    \end{align}
    where $W_{FF} \in \mathbb{R}^{1 \times \psi_z}$ is the weights matrix, $d_{z_H} \in \mathbb{R}^{\psi_z}$ is the final LSTM unit's hidden state, and $b_{FF} \in \mathbb{R}$ is the bias term. The output of the feed forward layer is $l_{FF} \in \mathbb{R}$.
    
    \item Sigmoid layer: this is the final layer of the classifier, and the output ($\hat{y}$) of this layer is obtained as described below:
    \begin{align}
        \hat{y} = \frac{1}{1 + e^{-l_{FF}}} = \sigma(l_{FF}) \label{eq:sigmoid},
    \end{align}
    where $\sigma(\cdot)$ denotes the sigmoid function. The sigmoid function ensures that $\hat{y}$ is in the range [0,1] and can be used as a probability estimate.
\end{itemize}

In the architecture described above, we have made multiple design choices including: (i) the use of LSTM units, and (ii) a single layer feed forward network after the LSTM layer. These choices are primarily motivated by simplicity and the prior success of LSTMs in a wide range of applications (\citealp{van2020review}).
The above architecture can be easily adapted to other RNN variants, as well as deeper feed forward networks after the LSTM layer. The sigmoid layer's output $\hat{y}$ is the probability estimate that the sequence $s$ has a positive label, and hence the training objective of the RNN classifier described above is to minimize
the binary-cross-entropy loss (\citealp{deep_learning_book_goodfellow}) as described below:
\begin{align} \label{eq:cross_entropy}
    loss_C = \;\; - \frac{1}{M} \sum_{s=1}^M  \left( y_s \log{\left( \hat{y}_s \right)}  + (1-y_s) \log{\left(1 - \hat{y}_s \right)}   \right),
\end{align}
where $loss_C$ needs to be minimized, $y_s$ is the binary label for the sequence $s$ ($1$ if positive, $0$ otherwise), $\hat{y}_s$ is the predicted probability of the sequence $s$ having a positive label (as in \Cref{eq:sigmoid}), and the sum is over $M$ training samples. 

%% RNN model %%
%\subsection{ML model for efficient ROV calculations} \label{sec:RNN_model}
%In this section, we cover the proposed RNN based ML model for efficient ROV calculations.

%can be implemented in a relatively short time compared to other deep-learning tools. \\

%% RNN training and evaluation %%
%\subsubsection{RNN model training and evaluation} \label{sec:RNN_model_train_eval}

%Given a set of $H$ zones with $H!$ sequences in set $S$ ($|S| = L = H!$), we randomly sample a small fraction of sequences ($frac_{seq}$) from $S$ to train the RNN model. Let $S_{m}$ denote the selected subset of sequences such that $|S_{m}|$ = $M$ ($M = frac_{seq}\times L$). On each of these $S_{m}$ sequences, we perform the ROV calculations using Algorithm 2 (described in \Cref{sec:RO_solution}). Once the policy values for $S_{m}$ sequences are obtained (denoted by $\eta_m$), the next step is to label these sequences with $1$s and $0$s to train the RNN sequence classifier. For this purpose, we use a binary labeling criteria as described below.\\

%% Labeling %%
\paragraph{Sequence labeling for RNN classifier} \label{sec:labeling}

The objective of labeling sequences in the training set is to feed the RNN model with positive and negative examples for sequence classification. The intuition around labeling sequences is as follows: sequences with label 1 should indicate that such type of zone orderings or patterns result in higher policy values compared to sequences with label 0.
The details of the proposed labeling processs are described below.
Let $\eta_{bin}$ denote a policy value threshold such that sequences with values higher than $\eta_{bin}$ are labeled as 1, or 0 otherwise.
%We define $\eta_{bin}$ based on the positive to negative sample ratio in the training set. In particular, we consider a maximum positive to negative ratio constraint $PNR_{max} \in [0,1]$ to regulate the number of positive and negative examples used in training the RNN model. By sorting the $\mathcal{S}_m$ train set sequences from highest to lowest policy values, we incrementally label $S'_m$ sequences as 1 and remaining $S_m -S'_m$ sequences as 0, and $\eta_{bin}$ is the threshold obtained with $\dfrac{S'_m}{S_m -S'_m}$ =  $PNR_{max}$. The objective is to analyze the performance sensitivity of the proposed ML model to different values of $PNR_{max}$ under multiple scenarios, to investigate the optimal range of $PNR_{max}$ for the RNN model to be able to effectively identify the complex sequence patterns in our setup. 
Assuming that the maximum of the policy values in the population sequences ($\mathcal{S}$) is known (let's say $\eta_{ub}$), one approach is to consider a value within a certain range of $\eta_{ub}$ to define $\eta_{bin}$. A previous study by \cite{chow2016reference} uses a sampling approach to approximate a lower bound for $\eta_{ub}$ from a subset of sequences. Assuming that the sequence policy values belong to an extreme value distribution (\emph{e.g.}, Weibull), they estimate the parameters from sampled sequences, and use those parameters to obtain the distribution of the maximum population value which is also Weibull distributed. We adopt a similar approach to estimate $\eta_{ub}$ in our setup. In particular, we fit a Weibull distribution to $\mathcal{S}_{m}$ policy values, and estimate the scale and shape parameters (\citealp{menon}). These are used to construct the distribution of the maximum for the population of all sequences $\mathcal{S}$ via a Weibull cumulative distribution function (CDF). The mean policy value at 50 percentile of the Weibull CDF is selected as $\eta_{ub}$. Using this, we calculate a policy threshold value ($\eta_{thr}$) as per \Cref{eq:ub_thrs}.
\begin{align}
    \eta_{thr} =  \eta_{ub} - thr_{fact} \times \eta_{ub}, \label{eq:ub_thrs}
\end{align}
where  $thr_{fact} \in [0,1]$ (\emph{i.e.}, a design choice).

One of the objectives of this study is to investigate the RNN model behavior to the changing number of training samples $M$. However, for the same service region scenario, the number of positively labeled sequences in $\mathcal{S}_m$ (using \Cref{eq:ub_thrs} for a given $thr_{fact}$) varies depending on the policy value distribution in the $M$ samples (which is mainly based on the configuration and characteristics of $H$ zones and the value of $M$). The ratio of positive and negative examples that is fed to the RNN model is a crucial part of the model learning process. Hence, to regulate the ratio of positive to negative samples in $\mathcal{S}_m$ sequences, we consider a maximum positive to negative ratio constraint $PNR_{max} \in [0,1]$. This is to ensure that the ratio of positive labels to negative labels obtained as per the above threshold for $M$ sequences ($\eta_{thr}$ in \Cref{eq:ub_thrs}) satisfy the $PNR_{max}$ constraint. By sorting the $\mathcal{S}_m$ training set sequences from highest to lowest policy values, we incrementally label $\mathcal{S}^{'}_m$ sequences (greater than $\eta_{thr}$) as 1 and the remaining $\mathcal{S}_m -\mathcal{S}^{'}_m$ sequences as 0, such that $\dfrac{\mathcal{S}'_m}{\mathcal{S}_m -\mathcal{S}^{'}_m}$ satisfy $PNR_{max}$; the resulting threshold policy value defines $\eta_{bin}$. The test set sequences are also labeled based on the  $\eta_{bin}$ obtained above. \\
%is defined as the threshold policy value obtained using such that $PNR_{max}$ is satisfied. 
%We conduct several experiments to analyze the performance sensitivity of the proposed ML model to different values of $PNR_{max}$ under multiple scenarios, to investigate what range of $PNR_{max}$ is optimal for the RNN model to be able to effectively identify the complex sequence patterns in our setup. 

%Using the $S_{m}$ labeled sequences, we then train the RNN sequence classifier. 
%

%% Labeling %%
\paragraph{RNN regressor as benchmark policy} \label{sec:blabeling}
 As a benchmark policy in our study, we consider an RNN based regression model. The regression model's training architecture is similar to the one in \Cref{fig:lstm_architecture} except changes in the last layer, and the loss function as explained below.
The sigmoid layer in Figure~\ref{fig:lstm_architecture} is replaced with a rectified linear unit (ReLU) layer and the output $\hat{y}$ is obtained using a standard ReLU activation function (\citealp{deep_learning_book_goodfellow}) as shown below:
\begin{align}
    \hat{y} = max(l_{FF}, 0). \label{eq:relu}
\end{align}
The training objective of the RNN regressor is to minimize the mean squared error loss ($loss_R$) (\citealp{deep_learning_book_goodfellow}) as shown in \Cref{eq:mse}.
\begin{align} \label{eq:mse}
    loss_R = \;\; - \frac{1}{M} \sum_{s=1}^M  ( y_s - \hat{y}_s)^{2},
\end{align}
where $y_s$ is the normalized policy value (explained below) for the sequence $s$, $\hat{y}_s$ is the predicted value of the sequence (as in \Cref{eq:relu}), and the sum is over $M$ training samples. The normalization\footnote{https://scikit-learn.org/stable/modules/generated/sklearn.preprocessing.StandardScaler.html} for $y_s$ is done on the basis of the mean and variance of policy values in the training data.

Therefore, the training process of the RNN regressor remains the same as RNN classifier, except that the normalized policy values of $\mathcal{S}_m$ sequences are used instead of labels to train the RNN regression model. The policy values predicted by the RNN regressor for $\mathcal{S}_r$ sequences are considered as the ranking scores, which is used to retrieve top $K$ sequences (out of $L-M$ sequences). Although the predicted  values by the regressor can directly be used as an estimate of the sequence policy values, we perform ROV calculations on the top $K$ sequences predicted by the RNN regressor for a fair comparison of the predictive abilities of the two models (\emph{i.e.}, RNN regressor and RNN classifier) using the evaluation metric as explained below.\\ 

%% Evaluation metric %%
\paragraph{RNN model evaluation}
Since the final output of the proposed RNN model (trained on $M$ samples) is the top $K$ predicted sequences retrieved by ranking the test set scores (\emph{i.e.}, $L-M$ sequences), we define an evaluation metric based on the recall metric commonly used in information retrieval and ranking setups (\citealp{inf_ret}). Recall at $K$ is the proportion of relevant items found in the top $K$ recommendations. When the actual number of relevant items is 1, then recall at $K$ is 1, if the model recommends that item in top $K$, else recall at $K$ is 0. To measure the performance of a trained RNN model on unseen ($L-M$) sequences, we select two sequences with the highest policy values from $\mathcal{S}_k$ and $\mathcal{S}_r$ respectively. The former gives the policy value of the best among the top $K$ sequences predicted by the ML model ($\eta_{pred}$), and the later provides the true value of the best sequence in $L-M$ test samples ($\eta_{true}$). Based on this, we define the following evaluation metric in our study:
\begin{align}
    Gap@K = \dfrac{\eta_{true} - \eta_{pred}}{\eta_{true}} \times 100\%, \label{eq:gap_k}
\end{align}
where $Gap@K$ represents the prediction gap between $\eta_{true}$ and $\eta_{pred}$ at top K. The lower the gap, the better is the performance of the model in predicting the top $K$ promising sequences out of $L-M$ samples.
The choice of this evaluation metric allows us to fairly compare different types of models (\emph{i.e.}, regression and classification), and to effectively measure sensitivity in the model performance (in terms of changes in the prediction gap) to varying input values.

%Additionally, we consider the metric widely used for evaluating binary classification models \emph{i.e.}, AUC (Area under the curve) ROC (Receiver Operating Characteristics) curve (\citealp{murphy}). In simple terms, the AUC score ($AUC \in [0,1]$) tells us how well the model is able to distinguish between positive and negative classes. 
 
Our objective is to investigate the effectiveness of the proposed CR-RNN policy in identifying the complex sequence patterns under multiple service region scenarios, and study the RNN model performance to varying input parameters in our setup to obtain useful insights. This includes number of zones $H$, zone volatilities $\sigma_H$, number of training samples $M$, $PNR_{max}$ for sequence labeling, and top $K$ sequences. For this, we conduct experiments on various combinations of input parameters as described in \Cref{sec:experiments} below.

%% Experiments %%
\section{Experiments} \label{sec:experiments}
\subsection{Experiment setup}
\begin{figure}[htp]%
\begin{center}
  \subcaptionbox{}{\includegraphics[width=3in]{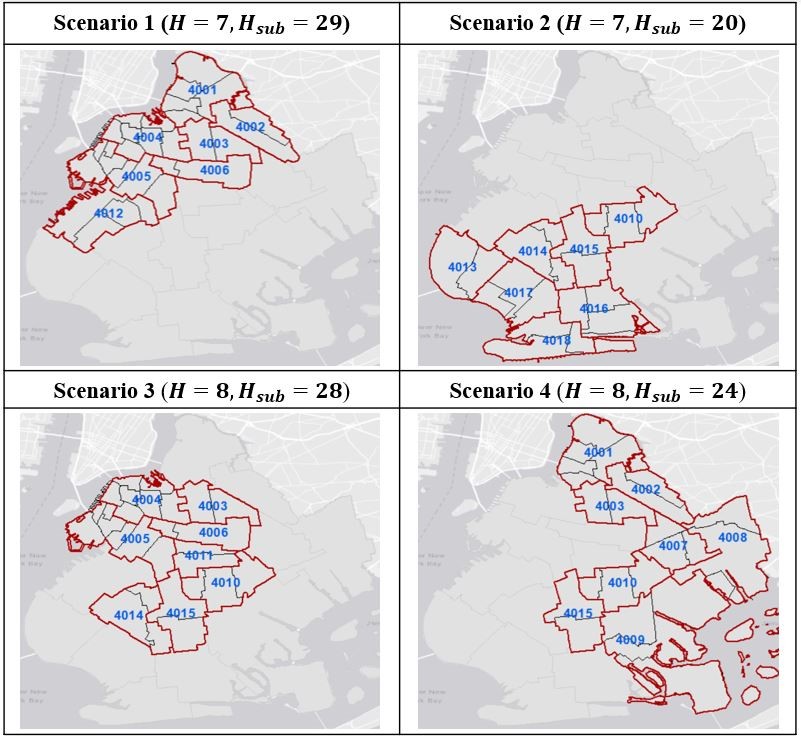}} \hspace{0.2em}%
  \subcaptionbox{}{\includegraphics[width=3in]{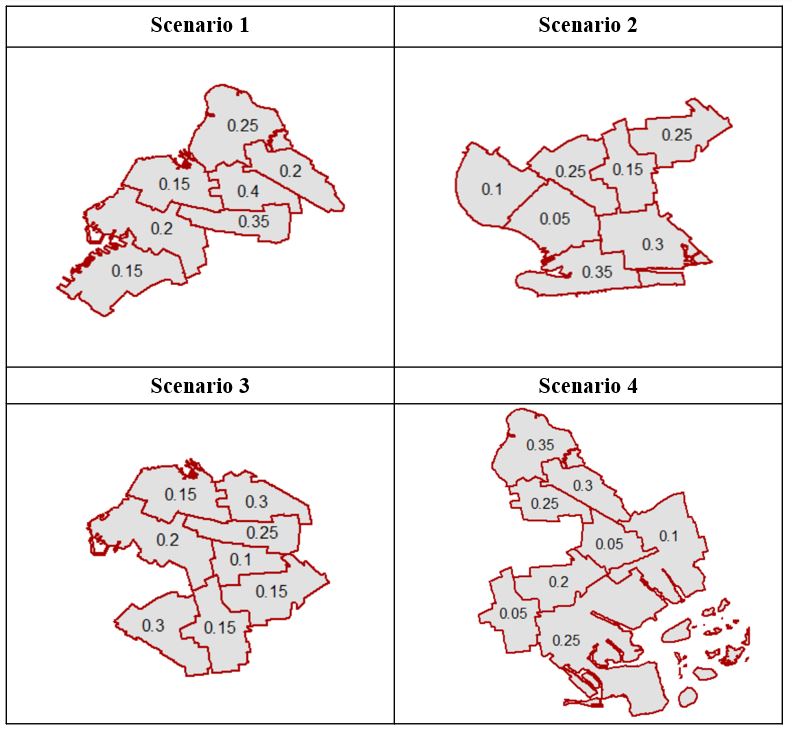}}
  \caption{Service region scenarios in NYC: (a) shows PUMA zones (service zones) along with zone IDs and PUMA covered taxi zones (sub-zones) highlighted in red and black outline respectively, (b) shows the service zone volatility values (defined in \Cref{sec:experiment_data}a).}
  \label{fig:scenarios}
\end{center}
\end{figure}

 Consider a MoD service provider planning to deploy their services to a region with $\mathcal{H}$ service zones and $\mathcal{H}_{sub}$ sub-zones, where each service zone $h \in \mathcal{H}$ constitute $h_{sub}$ sub-zones. As discussed in \Cref{sec:setup}, we assume the OD service demand in $\mathcal{H}_{sub}\times\mathcal{H}_{sub}$ sub-zones evolve independently as GBM motions with zero drift and hetereogeneous volatility (\emph{i.e.}, based on the zone volatilities $\sigma_H$). The objective is to determine the optimal investment strategy for the current time period $t_0$ (denotes a decision epoch) \emph{i.e.}, which service zones to defer and zones to invest in immediately; this is done by comparing the current NPV (under a deterministic setting) with the expanded NPV obtained using the RO model (considering service demand uncertainty for a $t_E$-year time horizon). For our experiments, we consider 4 different service region scenarios in Brooklyn (Kings County), NYC (with randomly selected geographically contiguous service zones) as shown in \Cref{fig:scenarios}(a). The $H$ service zones and $\mathcal{H}_{sub}$ sub-zones in each scenario are PUMA (Public Use Microdata Areas) zones and taxi zones in NYC\footnote{\label{nycopendata}opendata.cityofnewyork.us/} (each PUMA zone covers multiple taxi zones). Our experiments are carried out on a computer with Intel i7 processor with 2 cores, 4 logical processors and 16 GB RAM, and are implemented using Python (version 3.8.5). For our experiments (details in \Cref{sec:results}), we leverage TensorFlow (Keras\footnote{www.tensorflow.org}) for training the RNN models with the ADAM optimizer\footnote{keras.io/api/optimizers/adam}. Keras is popular across both industry and academic research communities as it provides a highly-flexible framework to implement deep neural networks, and has built-in optimizations to speed up training and inference.
 
%\Cref{sec:data} describes the demand data considered for the experiments along with the input parameters for the RO model and RNN model, \Cref{sec:ROmodel_results} includes the output from the RO model, and the results from the RNN model are discussed in \Cref{sec:RNNmodel_results}.

\subsection{Data} \label{sec:experiment_data}
We assume that the potential demand for the MoD services mainly comes from the auto and transit demand in the study area. The auto (drive alone and carpool) and public transit (bus, street car, subway, and rail) demand data for NYC is collected from Census Transportation Planning Products Program (\citealp{ctpp}), which provides mode-wise commute flows between census tracts. Based on the geographical boundaries of taxi zones and census tracts in NYC \footnote{opendata.cityofnewyork.us/}
%\footref{nycopendata} 
(each taxi zone covers multiple census tracts), we compute the aggregate commute flow across different taxi zones using the CTPP census tract demand data. We then calculate the peak hour commute flow as 23.38\% of the above travel demand; this value is based on the percentage auto and public transit flow during the morning peak hours (\emph{i.e.}, average of 7-8 AM and 8-9 AM values) in NYC obtained using American Community Survey (\citealp{acs}) data. Finally, we assume 60\% of the above demand as the potential (upper bound) OD demand ($\mathcal{H}_{sub}\times\mathcal{H}_{sub}$) for MoD services in our experiments. Although this study does not focus on the design of MoD services or investigation of travelers' modal preferences for such services, in practice, a mode choice model based on the socio-demographic, service operation characteristics, and other factors in the region should be considered to get a better estimation of the above demand values.
\bigskip
\paragraph{Parameters for ROV calculations in CR-RNN policy}
We assume the following for the ROV calculations.
  \begin{itemize}
  \setlength{\itemindent}{2.5em}
       \item Maturity time ($t_E$) = 5 years; $\mathcal{T}$ = \{1,2,3,4,5\}
        \item Number of basis functions ($J$) = 3
        \item Discount rate ($\rho$) = 2\% 
        \item Number of simulation paths ($|\mathcal{P}|$) = 300
        \item Value of time ($VoT$) = 0.293 euro/min (assuming \$20/hr as per prior studies (\citealp{holguin2012park, ma2022transit}))
        \item Customer perceived wait time factor ($\alpha_{W}$) = 2.1 (\citealp{kittelson2003transit})
        \item  Customer perceived in-vehicle time factor ($\alpha_{TIV}$) = 1 (\citealp{kittelson2003transit})
        \item Congestion scaling $\gamma$   = 0.005 (based on \citealp{wong_taxi})%range 0.002 - 0.04
        \item Vehicle speed ($v$)    = 19.31 km/hr (assuming 12 miles/hr average taxi speed in NYC (\citealp{nyctaxispeed}))
        \item Trip cost ($c_{ij}, i,j \in$ $\mathcal{H}_{sub}$)    = 2.42 euros (assuming a fixed price of \$2.75\footnote{new.mta.info/fares})
        \item Zone volatility values ($Z_{vol}$): \{5\%, 10\%, 15\%, 20\%, 25\%, 30\%, 35\%, 40\%\}; for each scenario with $H$ zones, we randomly assign volatility values from $Z_{vol}$ to define $\sigma_H$ (as shown in \Cref{fig:scenarios}(b)); therefore, as discussed earlier, for a zone $h$ with $\sigma_h$ volatility, all the OD pairs originating from the zone have $\sigma_h$ volatility. The OD demand volatility can be defined based on various other criteria as well, for example, specific OD pairs may be less volatile than others. Based on previous service implementations in similar regions or similar cities (\citealp{rath2022worldwide}) coupled with population and built environment data, service providers can define the the OD volatilities for the service region in consideration.
        \item Within-zone cost ($C_{wz}$): The ridership threshold that defines the within-zone cost for a set of service zones in our setup is considered to be 40\% of the average within-zone ridership (\emph{i.e.}, average of aggregate ridership within the $H$ PUMA zones in the region); this is  obtained based on the potential OD (taxi zone) demand at $t_0$ using Algorithm 1.
        \item Interzone cost ($C_{iz}$): The average interzone ridership (\emph{i.e.}, average of aggregate ridership between $H$ PUMA zones) is used to define the interzone cost for the service zones in our setup; this is obtained based on the potential OD (taxi zone) demand at $t_0$ using Algorithm 1.
    \end{itemize}
\bigskip

\paragraph{Parameters for RNN model}: We consider the following input parameters for the RNN model
\begin{itemize}
    \setlength{\itemindent}{2.5em}
    \item Training sample ratio ($frac_{seq}$): \{0.0012, 0.0028, 0.0044, 0.0068, 0.0084, 0.01, 0.02, 0.03, 0.04, 0.05, 0.06, 0.07, 0.08, 0.09, 0.1, 0.2, 0.3, 0.4, 0.5\}, where $frac_{seq}$ = 0.01 means 1\% of total population sequences ($L$) is used to train the RNN model.
    \item Top $K$ sequences ($K$): \{30, 50, 70\}
    \item Embedding sizes ($\psi_z$): \{10, 50, 100, 150, 200\}
    \item Positive to negative (training) sample ratio ($PNR_{max}$) = \{0.01, 0.02, 0.05\}, where $PNR_{max}$ = 0.01 denote 1\% positive to negative samples in the $M$ training set
\end{itemize}
%The model hyperparameters such as embedding size $\psi_z$ (for zone embeddings) and epochs are fine-tuned using a validation set. 
In the RNN model training process, we use 20\% of $\mathcal{S}_m$ sequences as a validation set for tuning hyper-parameters including epochs (upto 300) and embedding size (from the list above).
For the RNN classifier, we consider $thr_{fact}$ to be 0.1.
Although different values can be selected for the above mentioned hyper-parameters, to give a representative view of the results from our experiments, we consider hyper-parameter values as mentioned above. Study data and related code are included in \cite{figshare}\footnote{will be released upon publication}.

%Given the combinatorial complexity arising from multiple hyper-parameter combinations 
%and for better interpretation, we present the model results based on the above mentioned parameters (\emph{i.e.}, representative values chosen based on preliminary experiments). 

%Following this, using the trained RNN model, we test the model performance on remaining $L-M$ unseen sequences. For a particular $frac_{seq}$, we run the experiments three times to get the average prediction gap ($Gap@K$) for different $K$ values. 

%% Results %%
\section{Results and discussion} \label{sec:results}
This section first investigates the performance of the proposed CR-RNN policy (in comparison to CR policy without the RNN component) for service region design and timing decisions in multiple scenarios. We first discuss the CR policy results in \Cref{sec:RO_results}, followed by the CR-RNN model results in \Cref{sec:RNN_results} focusing on the overall reduction in the computational complexity. \Cref{sec:case study NYC} is a validation of the model using a case study in NYC, how the proposed CR-RNN policy can be used for efficient sequential service region design and timing decisions in the context of expansion of service region to new areas and share useful insights. 

\subsection{CR policy results} \label{sec:RO_results}
We apply the CR policy (\emph{i.e.}, ROV calculations without the RNN model in \Cref{alg:cr_rnn_algo}) to each of the four service region scenarios in NYC (see \Cref{fig:scenarios}) to determine the optimal investment strategies for the service zones in respective scenarios. The number of ordered sequences for scenarios 1 and 2 (with $H$ = 7 service zones each) is 5040, and that for scenarios 3 and 4 (with $H$ = 8 service zones each) is 40,320. \Cref{tab:scenario_seqvals} summarizes the ridership threshold values (\emph{i.e.}, within-zone cost $C_{wz}$ and interzone cost $C_{iz}$) and the CR policy output for each scenario including: NPV (\emph{i.e.}, total payoff from investment in all $H$ zones under a deterministic setting in the current time period $t_0$),  highest and lowest option values (\emph{i.e.}, expanded NPV including deferral premium), and the respective sequences among $H!$ sequences. For each scenario, \Cref{tab:scenario_seqvals} shows the plot of the option value in each ordered sequence (\emph{i.e.}, value of the first zone (project) in the sequence). 

\begin{figure}[!htb]
\begin{center}
\includegraphics[scale=.70]{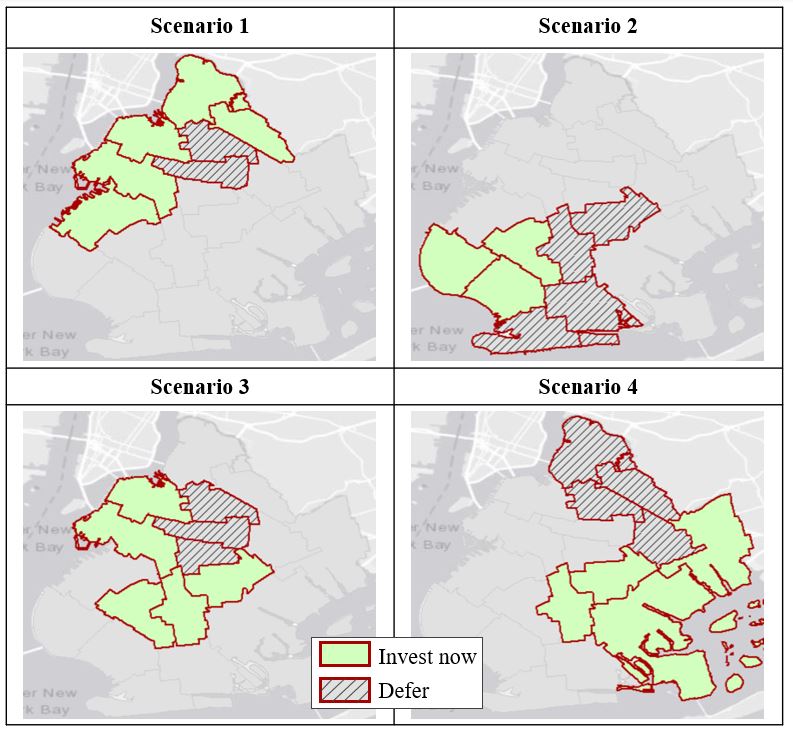}
\caption{Optimal investment strategy for service zones (\emph{i.e.}, invest immediately or defer investment) obtained based on the CR policy for the service region scenarios.} 
\label{fig:RO_strategies} 
\end{center}
\end{figure}

The optimal investment strategies obtained for each scenario are illustrated in \Cref{fig:RO_strategies}, where the model suggests investing immediately in selected zones (as highlighted in the figure) for the MoD service region design, while deferring the investment of the remaining zones for later. The experiments suggest that, compared to investing immediately in all the zones based on NPV (column 4 in \Cref{tab:scenario_seqvals}), there is value in waiting to invest in some zones using the RO approach (column 6 of \Cref{tab:scenario_seqvals}). Although the magnitude of this value depends on the payoff function, zone volatilities among other parameters; the consistency in the relative high value of the wait-and-see option across different scenarios highlights the benefits of the RO approach in service region investment decisions under demand uncertainty. 

%We discuss below the results of the proposed RNN model, focusing on the reduction in the relative computational cost with the use of an ML approach in RO model calculations. 

\begin{table}[h!]
  \centering
  \begin{tabular}{ | >{\centering\arraybackslash}p{0.8cm} | c | c |c |>{\centering\arraybackslash}p{2cm} | >{\centering\arraybackslash}p{1cm} |>{\centering\arraybackslash}p{2cm} | >{\centering\arraybackslash}p{1cm}| >{\centering\arraybackslash}p{0.7cm}| c|}
    \hline
    Scenario & $C_{wz}$ & $C_{iz}$ & NPV & Sequence with & Option & Sequence with & Option & Run & Sequence policy value  \\ 
     &   &   &  & highest option value & value (highest) & lowest option value & value (lowest) & time  & distribution \\
    \hline
    1  & 49 & 68 & 516 & \{4001, 4002, 4005, 4004, 4012, 4006, 4003\} & 1025.17 & \{4006, 4003, 4004, 4012, 4002, 4001, 4005\} & 738.4 & 0.69 hr
    & \begin{minipage}{.18\textwidth}
      \includegraphics[width=\linewidth, height=22mm]{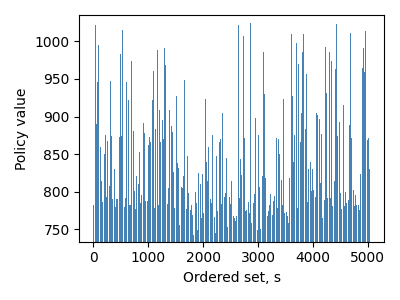}
    \end{minipage}
    \\ \hline
    2  & 65 & 59 & 684 & \{4017, 4013, 4014, 4016, 4018, 4010, 4015\} & 1005.05  & \{4016, 4018, 4015, 4017, 4010, 4013, 4014\} & 800.97 & 0.77 hr
    & \begin{minipage}{.18\textwidth}
      \includegraphics[width=\linewidth, height=22mm]{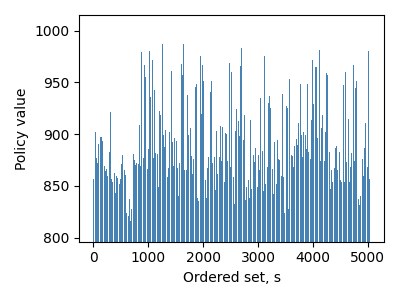}
    \end{minipage}
    \\ \hline
    3 & 51 & 61 & 613 & \{4010, 4005, 4004, 4015, 4014, 4003, 4006, 4011\} & 1061.19  & \{4003, 4014, 4010, 4015, 4005, 4011, 4006, 4004\}& 759.86 & 4.17 hrs
    & \begin{minipage}{.18\textwidth}
      \includegraphics[width=\linewidth, height=22mm]{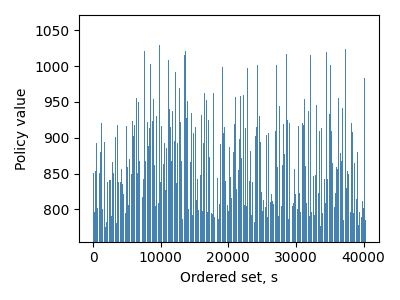}
    \end{minipage}
    \\ \hline
    4 & 45 & 44 & 538 & \{4010, 4008, 4009, 4015, 4003, 4007, 4002, 4001\} & 801.07  & \{4002, 4001, 4003, 4007, 4015, 4008, 4010, 4009\} & 652.88 & 4.32 hrs
    & \begin{minipage}{.18\textwidth}
      \includegraphics[width=\linewidth, height=22mm]{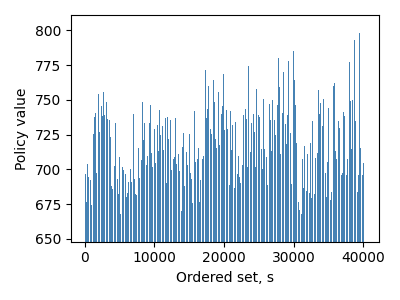}
    \end{minipage}
    \\ \hline
    
  \end{tabular}
  \caption{Scenario-wise input parameters and CR policy values.}\label{tab:scenario_seqvals}
\end{table}

The optimal investment decisions (as shown in \Cref{fig:RO_strategies}) correspond to the optimal sequence (\emph{i.e.}, one with the highest option value, see \Cref{tab:scenario_seqvals}). On the contrary, the policy with the lowest option value suggests investing in one zone for scenario 1 (\emph{i.e.,} zoneID 4006), one zone in scenario 3 (\emph{i.e.}, zoneID 4003), deferring investment in all zones in scenario 2, and investing in two zones in scenario 4 (\emph{i.e.}, zoneIDs 4001 and 4002). In each of these cases, the sequence policy values (as shown in \Cref{tab:scenario_seqvals}) are 18-28\% lower than the ones obtained using the optimal sequences for the scenarios. This shows the impact the zone orderings can have in the value of the option and the investment decisions. The total computational time involved in the ROV calculations depends on the number of simulation paths, time steps and payoff function considered; based on the assumed values in our setup, it is observed that, when the number of service zones in the region increases from 7 to 8, the average computation time increases about 6 times (see \Cref{tab:scenario_seqvals}). Since the proposed CR-RNN policy focuses on the relative reduction in the overall computational cost with the use of an ML approach, the benefits of the proposed framework holds true (with varying degrees) for similar setups involving RO applications.

\subsection{CR-RNN policy model results} \label{sec:RNN_results}
To study the effectiveness of the LSTM based RNN model in predicting the sequence policy values, we conduct experiments with different combinations of training sample ratio ($frac_{seq}$), $PNR_{max}$ and top $K$ sequences. In particular, we explore 57 individual RNN sequence classification models (19 training ratios $\times$ 3 $PNR_{max}$ choices), and 19 individual RNN regression models. Furthermore, each of the LSTM classification and regression models are trained and tested 3 times with different random splits (to get a better estimation of the model performance), and the average model prediction gap $Gap@K$ is computed for the test set for $K \in \{30, 50, 70\}$ . The model learning rate and batch size are set to 0.001 and 32 respectively, besides tuning the epochs and embedding size (using a validation set as discussed in \Cref{sec:experiment_data}) in the model training process. 

\begin{figure}[htp]%
\begin{center}
  \subcaptionbox{}{\includegraphics[width=5in, height=3.05cm]{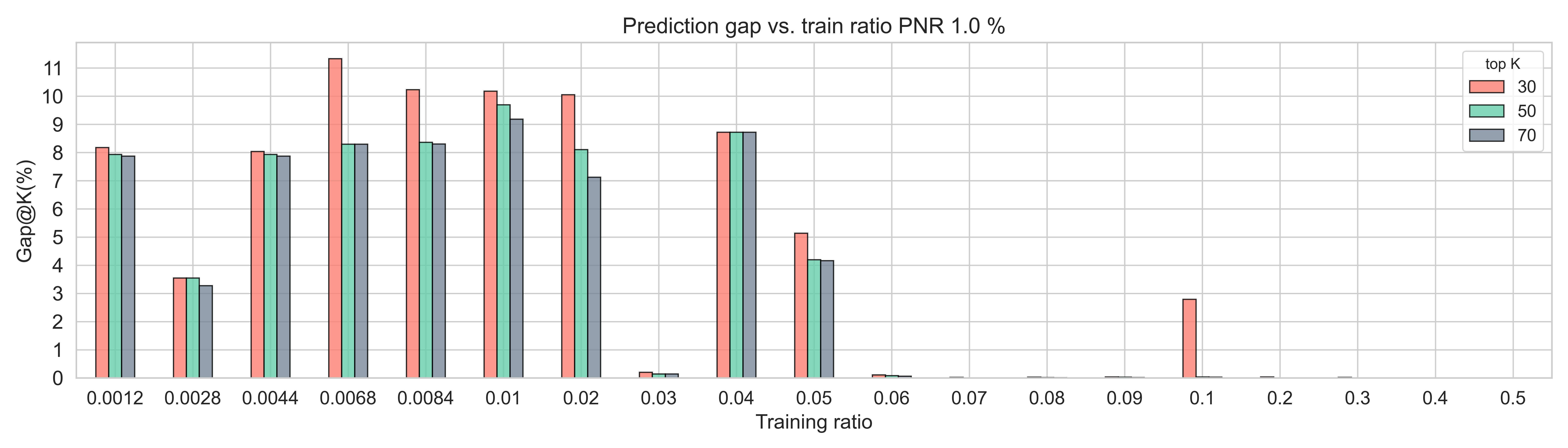}} 
  \subcaptionbox{}{\includegraphics[width=5in, height=3.05cm]{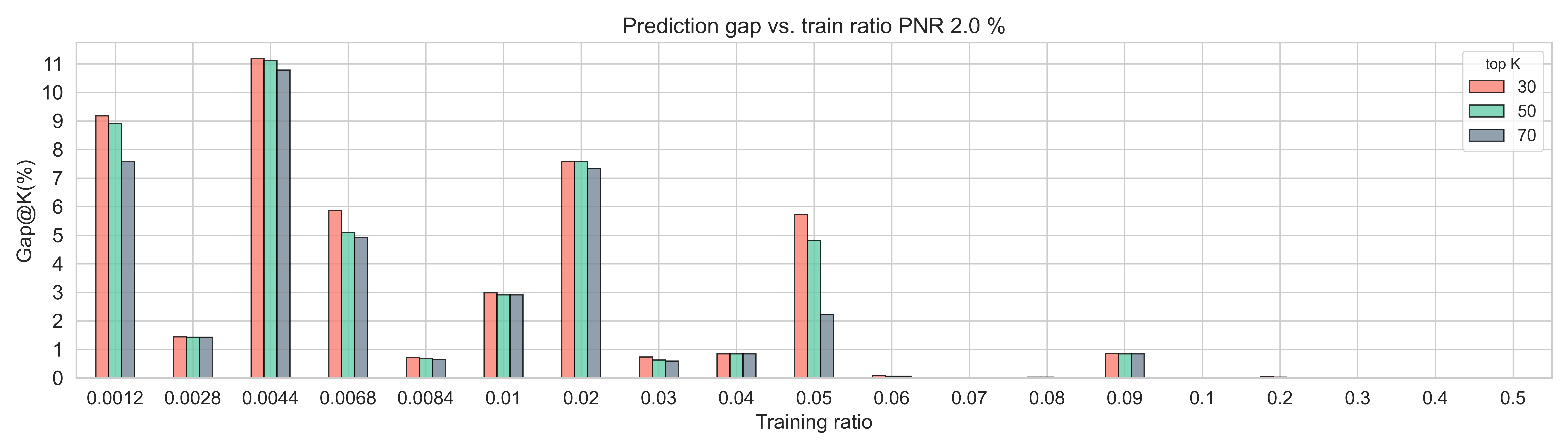}}
  \subcaptionbox{}{\includegraphics[width=5in, height=3.05cm]{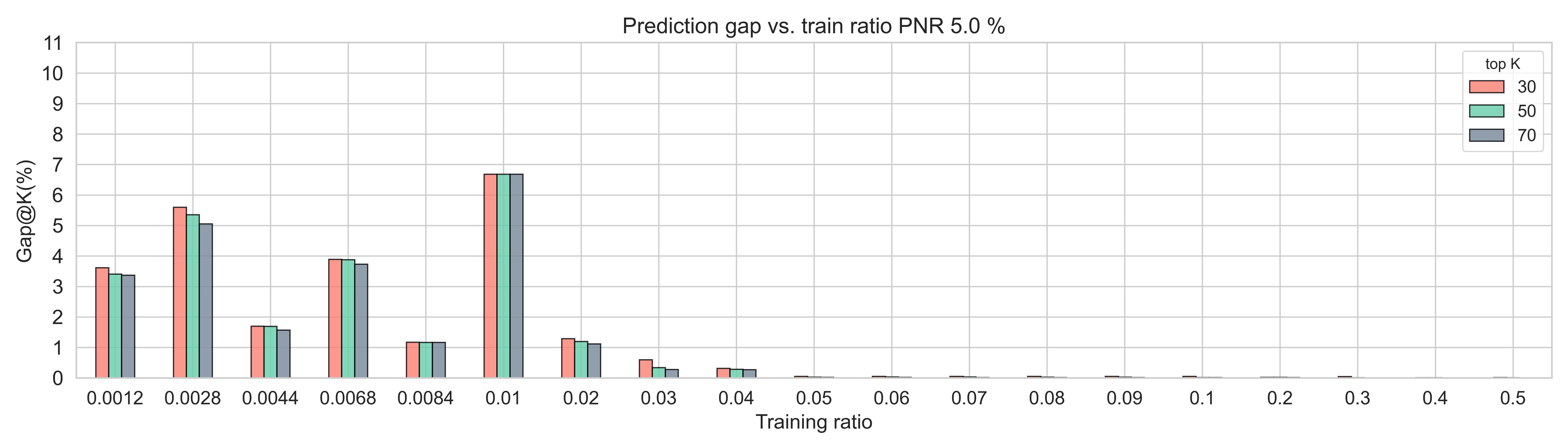}}
  \caption{Scenario 1 (7 zones) RNN classification model prediction gap ($Gap@K$) for different training ratios for each $K$ in top $K$ (in legend) with (a) $PNR_{max}$ = 1\%, (b) $PNR_{max}$ = 2\%, (c) $PNR_{max}$ = 5\%.}
  \label{fig:RNN_cl_results_sc1}
\end{center}
\end{figure}

\begin{figure}[htp]%
\begin{center}
  \subcaptionbox{}{\includegraphics[width=5in, height=3.05cm]{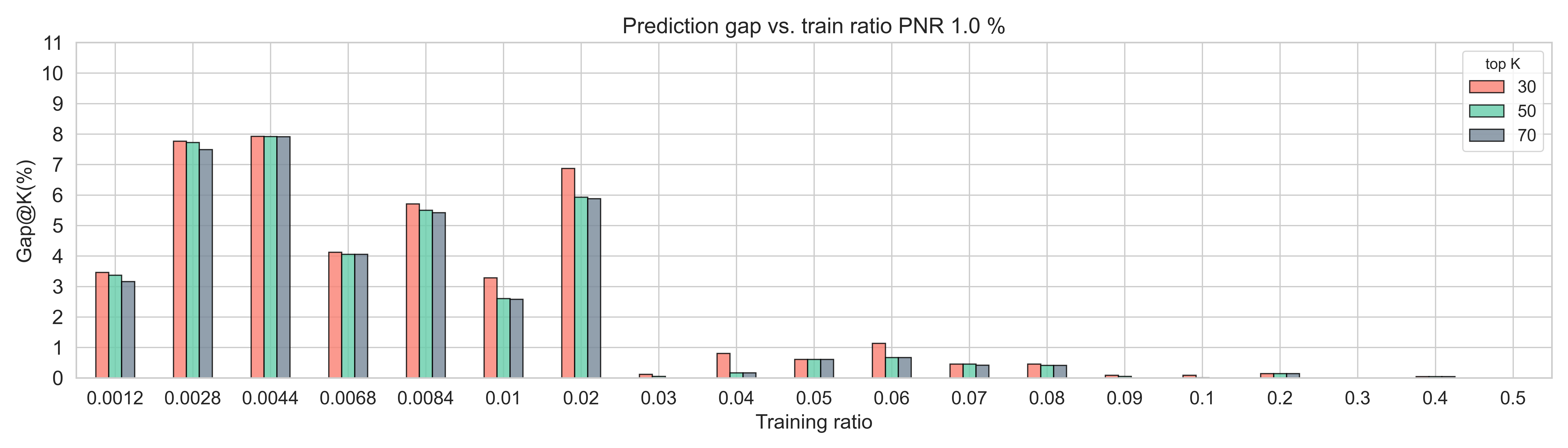}}
  \subcaptionbox{}{\includegraphics[width=5in, height=3.05cm]{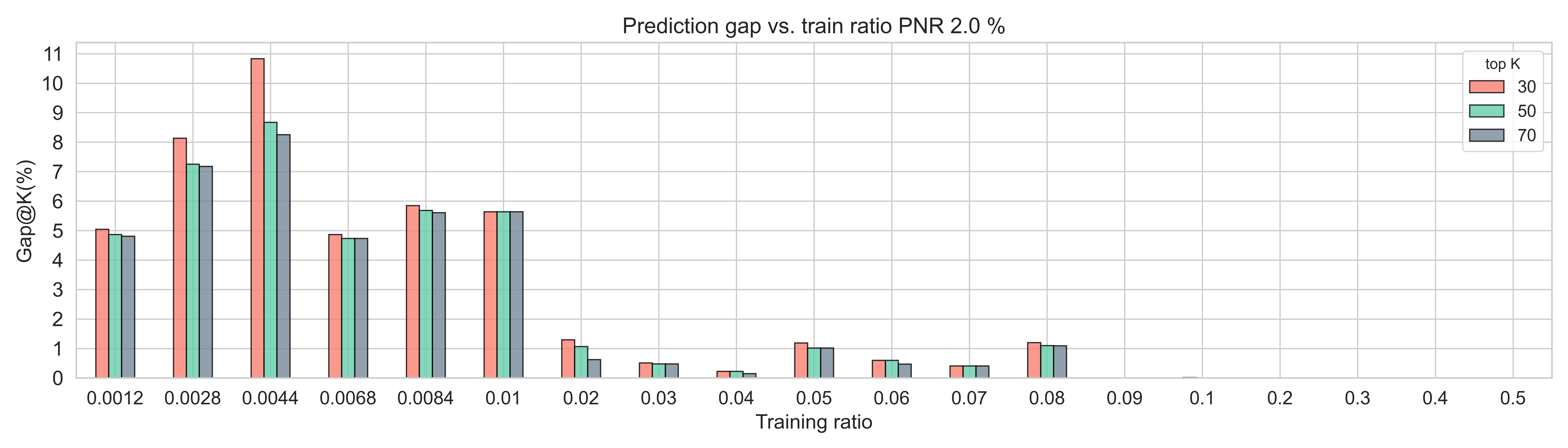}}
  \subcaptionbox{}{\includegraphics[width=5in, height=3.05cm]{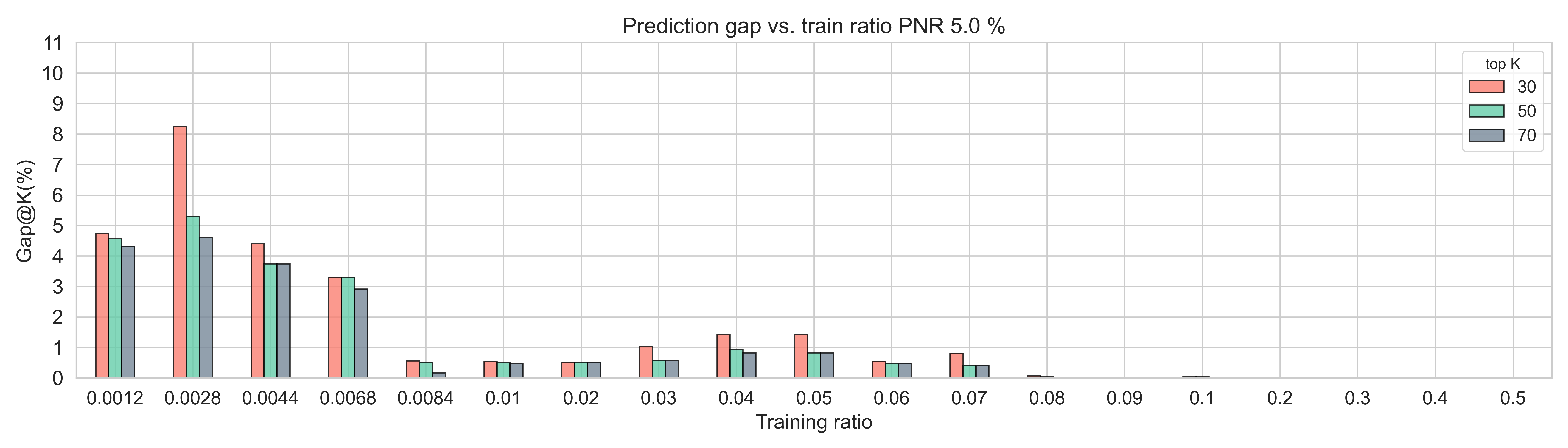}}
  \caption{Scenario 2 (7 zones) RNN classification model prediction gap ($Gap@K$) for different training ratios for each $K$ in top $K$ (in legend) with (a) $PNR_{max}$ = 1\%, (b) $PNR_{max}$ = 2\%, (c) $PNR_{max}$ = 5\%.}
  \label{fig:RNN_cl_results_sc2}
\end{center}
\end{figure}

\begin{figure}[htp]%
\begin{center}
  \subcaptionbox{}{\includegraphics[width=5in, height=3.05cm]{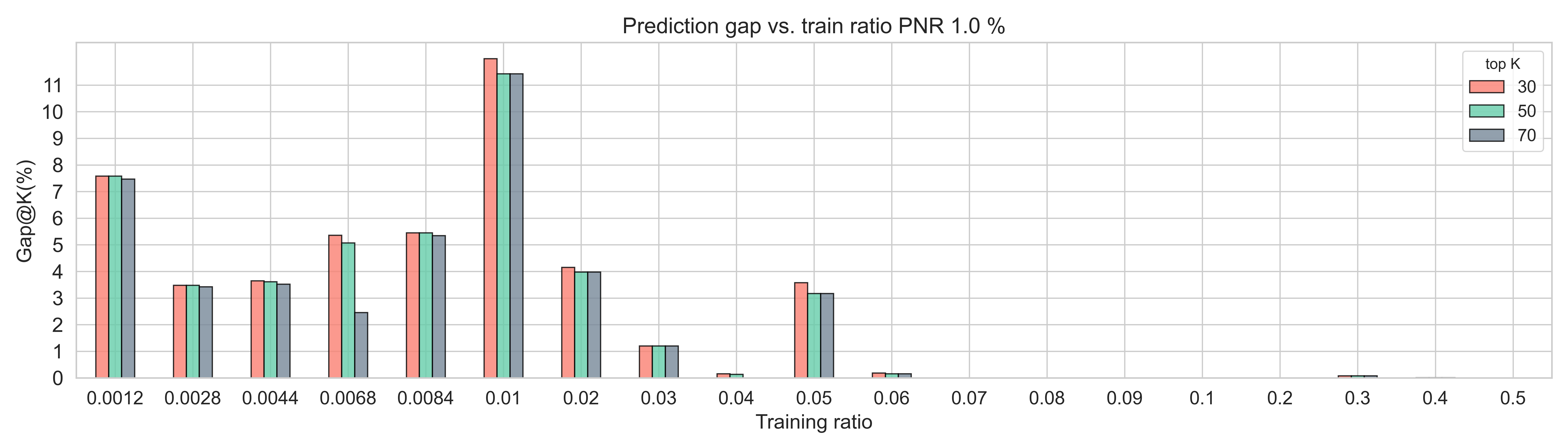}} 
  \subcaptionbox{}{\includegraphics[width=5in, height=3.05cm]{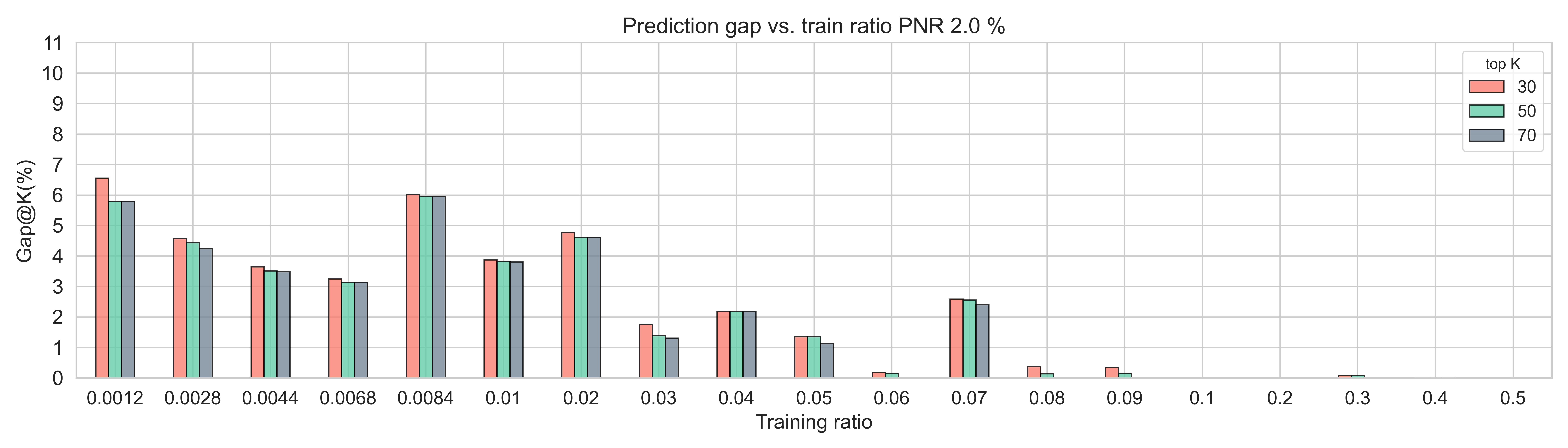}}
  \subcaptionbox{}{\includegraphics[width=5in, height=3.05cm]{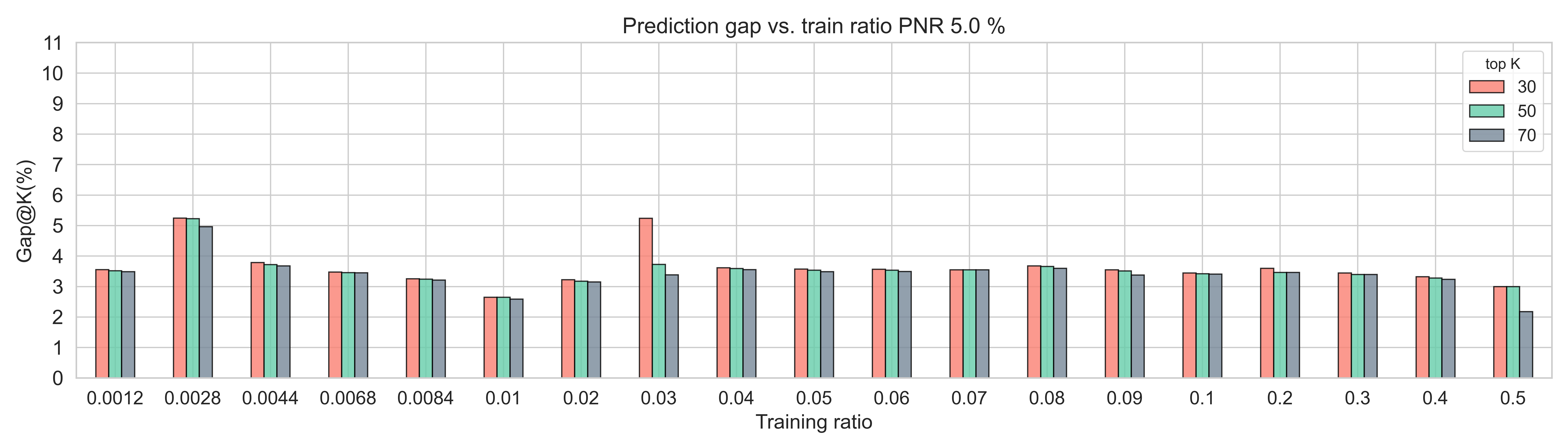}}
  \caption{Scenario 3 (8 zones) RNN classification model prediction gap ($Gap@K$) for different training ratios for each $K$ in top $K$ (in legend) with (a) $PNR_{max}$ = 1\%, (b) $PNR_{max}$ = 2\%, (c) $PNR_{max}$ = 5\%.}
  \label{fig:RNN_cl_results_sc3}
\end{center}
\end{figure}

\begin{figure}[htp]%
\begin{center}
  \subcaptionbox{}{\includegraphics[width=5in, height=3.2cm]{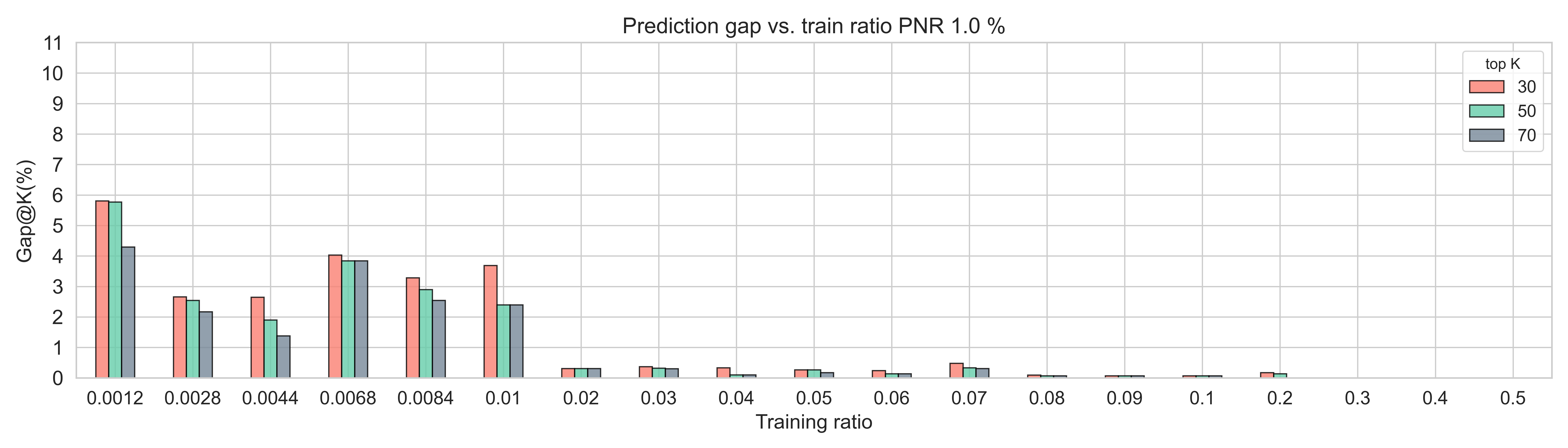}} 
  \subcaptionbox{}{\includegraphics[width=5in, height=3.2cm]{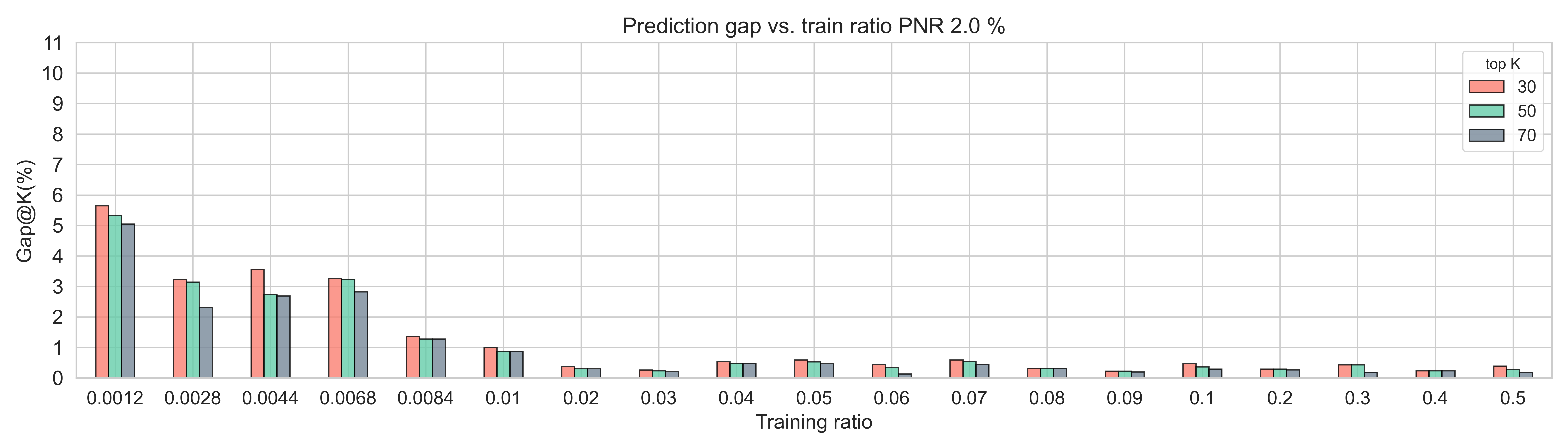}}
  \subcaptionbox{}{\includegraphics[width=5in, height=3.2cm]{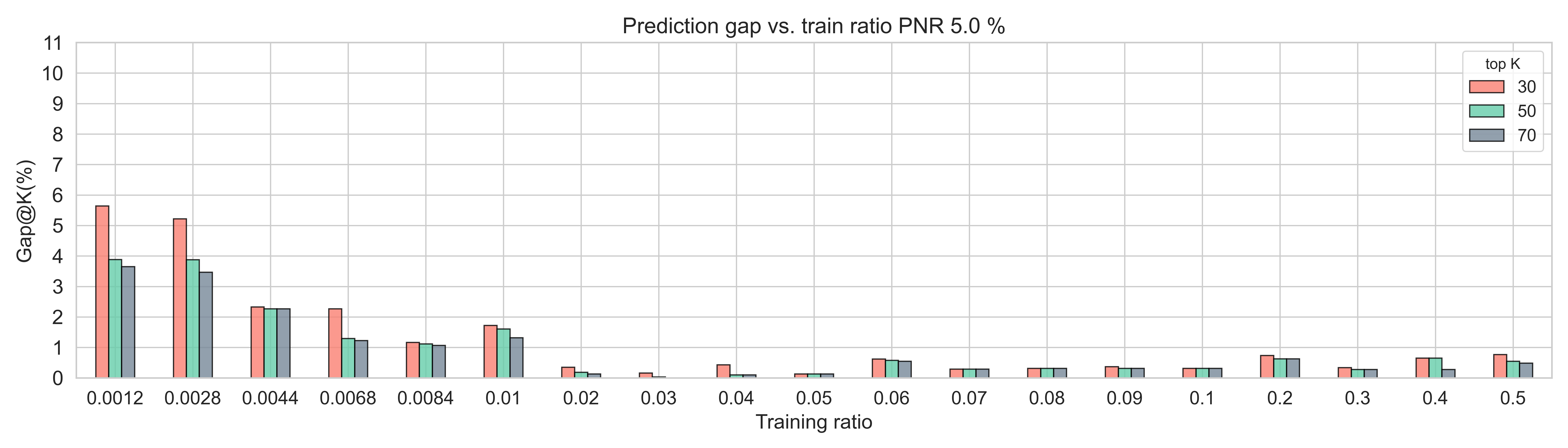}}
  \caption{Scenario 4 (8 zones) RNN classification model prediction gap ($Gap@K$) for different training ratios for each $K$ in top $K$ (in legend) with (a) $PNR_{max}$ = 1\%, (b) $PNR_{max}$ = 2\%, (c) $PNR_{max}$ = 5\%.}
  \label{fig:RNN_cl_results_sc4}
\end{center}
\end{figure}

 \Cref{fig:RNN_cl_results_sc1,fig:RNN_cl_results_sc2,fig:RNN_cl_results_sc3,fig:RNN_cl_results_sc4} illustrate the RNN classifier performance for the four scenarios (for each $PNR_{max}$). The plots represent $Gap@K$ for each training ratio $frac_{seq}$ for different $K$ values. Across all the four scenarios, for $PNR_{max}$ 0.01 and 0.02, the prediction gap monotonically decreases when training samples are increased. This is important, since it highlights the classifier's ability to properly learn the complex sequence patterns, which is further improved when more examples are fed into the training process. Moreover, in some scenarios (\emph{e.g.}, scenario 3 with 8 zones), it is seen that increasing the $PNR_{max}$ (\emph{e.g.}, 0.05) decreases the model performance. This is because, for a higher $PNR_{max}$, the number of positive samples increases due to a lower policy threshold value $\eta_{bin}$; this may not necessarily be beneficial to the model learning process and instead may be providing bad examples of promising sequences to the model. Moreover, for the above cases, it is noticed that the sequence labeling is mainly based on the $PNR_{max}$ value, and using $\eta_{thr}$ (obtained as per the assumed $thr_{fact}$) is not as effective in the model training process (since it mostly generated higher positive to negative ratio than $PNR_{max}$ values considered). Hence, the results presented are similar to using only $PNR_{max}$ for labeling purpose. 
 
\begin{figure}[htp]%
\begin{center}
  \subcaptionbox{}{\includegraphics[width=5in, height=3.05cm]{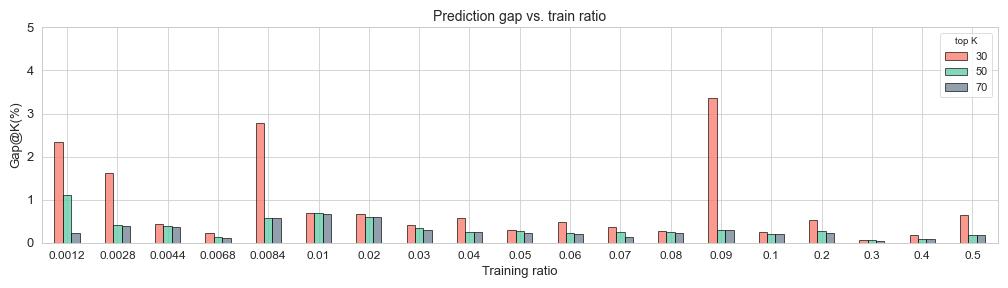}} 
  \subcaptionbox{}{\includegraphics[width=5in, height=3.05cm]{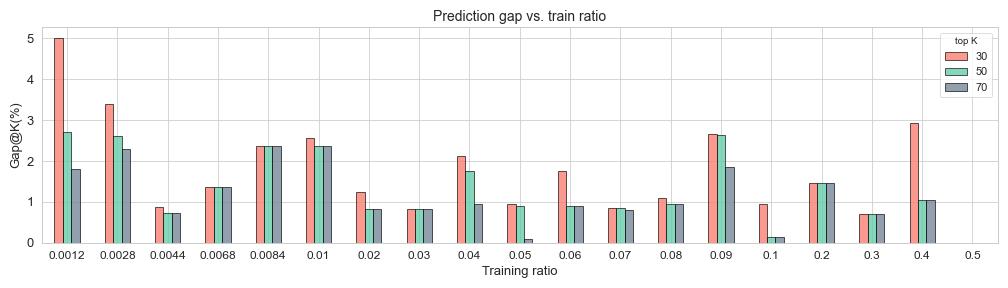}}
  \subcaptionbox{}{\includegraphics[width=5in, height=3.05cm]{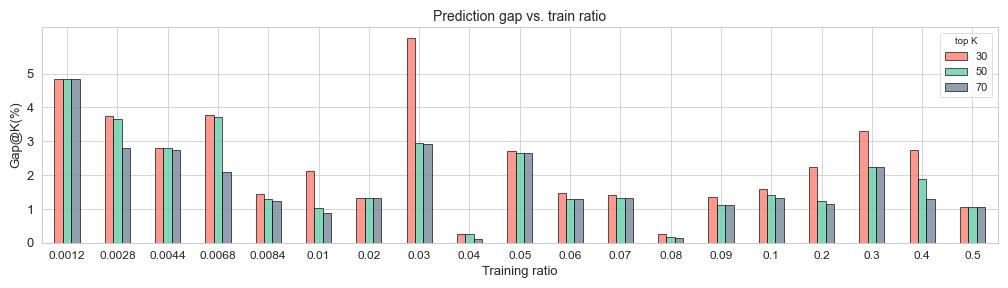}} 
  \subcaptionbox{}{\includegraphics[width=5in, height=3.05cm]{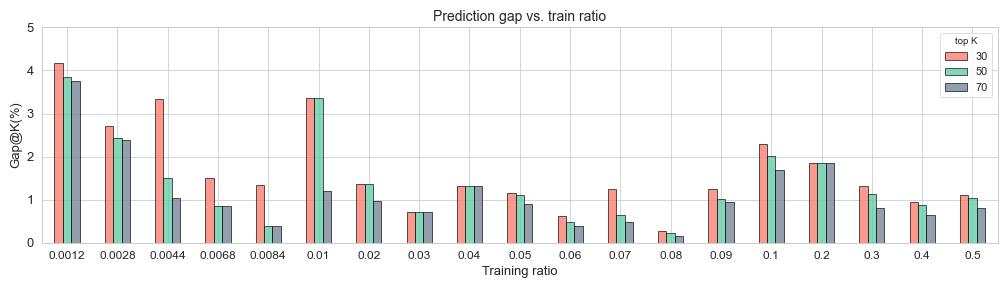}}
  \caption{RNN regression model prediction gap ($Gap@K$) for different training ratios for each $K$ in top $K$ (in legend) for scenarios with 7 zones: (a) Scenario 1 and (b) Scenario 2, and scenarios with 8 zones: (c) Scenario 3 and (d)  Scenario 4.}
  \label{fig:RNN_reg_results}
\end{center}
\end{figure}

 On the other hand, \Cref{fig:RNN_reg_results} illustrates the RNN regression model performance on the four scenarios. For scenario 1 (with 7 zones), it is observed that the average gap lies within 1\% for $K \geq 50$ with $frac_{seq} \geq 0.0028$. However, this behavior is not consistent across the other three scenarios, where the gap values are relatively higher, even with increased $frac_{seq}$ values. Moreover, as the figure shows, for most scenarios, there is no decreasing trend observed in the model prediction gap with increment in the training samples for the $K$ values considered, and more training samples ($\geq$ 50\% of total sequences) may be required for the model to achieve better prediction accuracy. 
 
Therefore, based on the experiment results across multiple scenarios in our setup, we can infer the following. The proposed method of selecting promising candidate sequences for policy evaluation in sequential service region design problem using an RNN classification model performs better than the benchmark RNN regression model.  Moreover, despite the sequences being generated from the same set of zones, the superiority of RNN classifier over RNN regressor on the same scenarios using the same training data highlights the relevance and ability of RNN classification model to effectively learn and capture the complex investment sequence patterns in our setup. A relatively small $PNR_{max}$ (about 1-2\%) with a minimum 6-8\% sample sequences is sufficient for the RNN classifier to capture useful service zone patterns and sub-sequence orderings. Such a classifier is able to effectively predict top $K$ promising zone sequences (for $K \geq$ 50) from the new unseen sequences with an average prediction gap of less than 1\% in most cases.
Including the top $K$ = 50 model predicted sequences, for scenarios with 7 service zones, 6\% training ratio is equal to 352 sequences (out of 5040 population sequences), and that for scenarios with 8 service zones is 2469 sequences (out of 40,320 population sequences). In these cases, it is worth highlighting the effectiveness of the RNN classifier in predicting the top 50 promising sequences out of 4738 sequences and 37,901 sequences for scenarios with 7 and 8 service zones respectively. 
%The best performing models observed in our setup are for $PNR_{max}$ in the range 1-2\%, hence further investigation on the effect of lower $thr_{fact}$ values is not done in the study. However, the use of $\eta_{thr}$ along with $PNR_{max}$ for different setups, as well as other labeling strategies to further improve the RNN model performance could be explored in the future.

\newpage
\Cref{tab:model_time} summarizes the average time (for each set of service zones) used in training the RNN classifiers 
%(with different $frac_{seq}$ values for $PNR_{max}$ = 1\%) 
along with the prediction time (for predicting the probability scores for the test set sequences of size $L-M$) for different training ratios. Additionally, the table reports the metric widely used for evaluating binary classification models \emph{i.e.}, AUC (Area under the curve) ROC (Receiver Operating Characteristics) curve (\citealp{murphy}). In simple terms, the AUC score ($AUC \in [0,1]$) tells us how well the model is able to distinguish between positive and negative classes. The values reported in \cref{tab:model_time} for 7 zones and 8 zones are averaged across scenarios 1 and 2, and scenarios 3 and 4 respectively. The high test AUC values and the short training and prediction time demonstrate the efficiency of the proposed RNN classification models. 
%ML approach by substantially reducing the computational complexity of the RO model. 
Therefore, by using the proposed ML approach in our setup, we achieve a substantial reduction in the overall computational cost (\emph{i.e.}, reduction in the computation time for more than 90\% of investment sequences).

\begin{table}[h!]
  \centering
  \renewcommand{\arraystretch}{1.3}
  \begin{tabular}{ | p{0.7cm}|p{1.1cm} | p{0.55cm}|p{0.55cm}| p{0.55cm}|  p{0.55cm} | p{0.55cm} | p{0.45cm}|p{0.45cm} | p{0.45cm}|  p{0.45cm} | p{0.45cm} | p{0.45cm} |p{0.45cm} |p{0.45cm} |p{0.45cm} |p{0.45cm}|}
    \hline
Service 	&	Training 	&	0.12\%	&	0.28\%	&	0.44\%	&	0.68\%	&	0.84\%	&	1\%	&	2\%	& 3\%	&	4\%	&	5\%	&	6\%	&	7\%	&	8\%	&	9\%	&	10\%\\ 
zones	& ratio (\%)	&		&		&		&	&		&		&		& 	&		&		&		&		&		&		&	\\ \hline

\multirow{3}{*}{7 zones} &		train time	&	2.42	&	2.15	&	2.35	&	2.76	&	2.58	&	2.26	&	3.32	&	3.84	&	2.89	&	5.16	&	5.11	&	5.5	&	6.87	&	9.2	&	7.63 \\ \cline{2-17}
&	test time	&	0.56	&	0.55	&	0.62	&	0.6	&	0.6	&	0.56	&	0.58	&	0.55	&	0.67	&	0.66	&	0.62	&	0.61	&	0.57	&	0.56	&	0.61\\ \cline{2-17}
	&	AUC	&	0.71	&	0.67	&	0.8	&	0.78	&	0.83	&	0.81	&	0.88	&	0.92	&	0.92	&	0.96	&	0.96	&	0.98	&	0.98	&	0.99	&	0.99\\ \hline
	
\multirow{3}{*}{8 zones} &	train time	&	2.95	&	3.82	&	4.4	&	6.24	&	6.4	&	5.2	&	11.7	&	13.74	&	13.78	&	13.96	&	22.22	&	20.56	&	22.24	&	26.52	&	26.94\\ \cline{2-17}
	&	test time	&	2.93	&	2.82	&	3.14	&	3.0	&	3.36	&	2.87	&	2.86	&	2.69	&	2.74	&	2.9	&	3.0	&	2.64	&	3.0	&	2.9	&	2.48\\ \cline{2-17}
	&	AUC	&	0.8	&	0.8	&	0.86	&	0.92	&	0.91	&	0.88	&	0.98	&	0.99	&	0.99	&	0.99	&	0.99	&	0.99	&	0.99	&	0.99	&	0.99\\ \hline
  \end{tabular}
  \caption{Average RNN classification model training time and testing time (in seconds) and test set AUC scores for different training ratio ($frac_{seq}$ in \%) with $PNR_{max}$ = 1\%.}\label{tab:model_time}
\end{table}

\newpage
Additionally, for higher number of service zones, further reduction in computation time can be obtained from the SEQ component (refer \Cref{fig:ROV_components}). In particular, mobility operators or planners can prune out sequences with specific patterns that are operationally less feasible \emph{i.e.},  only such subset sequences can be enumerated (out of $H!$ sequences) in which each possible continuous sub-sequence constitute only feasible zones \emph{i.e.}, zones fulfilling specific criteria (\emph{e.g.}, within a maximum radius in terms of average travel distance or travel time) based on the operation strategies. For example, consider a scenario with a region constituting 5 candidate service zones ($z_1, z_2, z_3, z_4$ and $z_5$, such that $H = 5$) and assume we have an operation threshold for maximum average travel time in the service region as $TT_{max}$. In this case, we can prune out sequences for which the average travel time for any continuous sub-sequence (group of zones) starting from the first zone in a sequence exceeds this value. Essentially, while enumerating a sequence \emph{e.g.}, $s_1 = \{z_1, z_2, z_4, z_3, z_5\}$, we iteratively check if the following sub-sequences fulfill the given criteria/threshold (i) $z_1$ and $z_2$, (ii) $z_1$,  $z_2$, and $z_4$, (iii) $z_1$,  $z_2$, $z_4$, and $z_3$, and so on. If, for example (ii) exceeds the threshold, we stop the iteration and prune the sequence $s_1$ from $H!$. Therefore, the choice of different threshold values for the defined criterion would yield different length subset sequences ($\leq H!$); such scaled down subset sequences can further improve the efficiency of the proposed CR-RNN policy.  

\newpage
\subsection{Validation case study: MoD service area expansion in NYC} \label{sec:case study NYC}
\begin{figure}[!htb]
\begin{center}
\includegraphics[scale=.47]{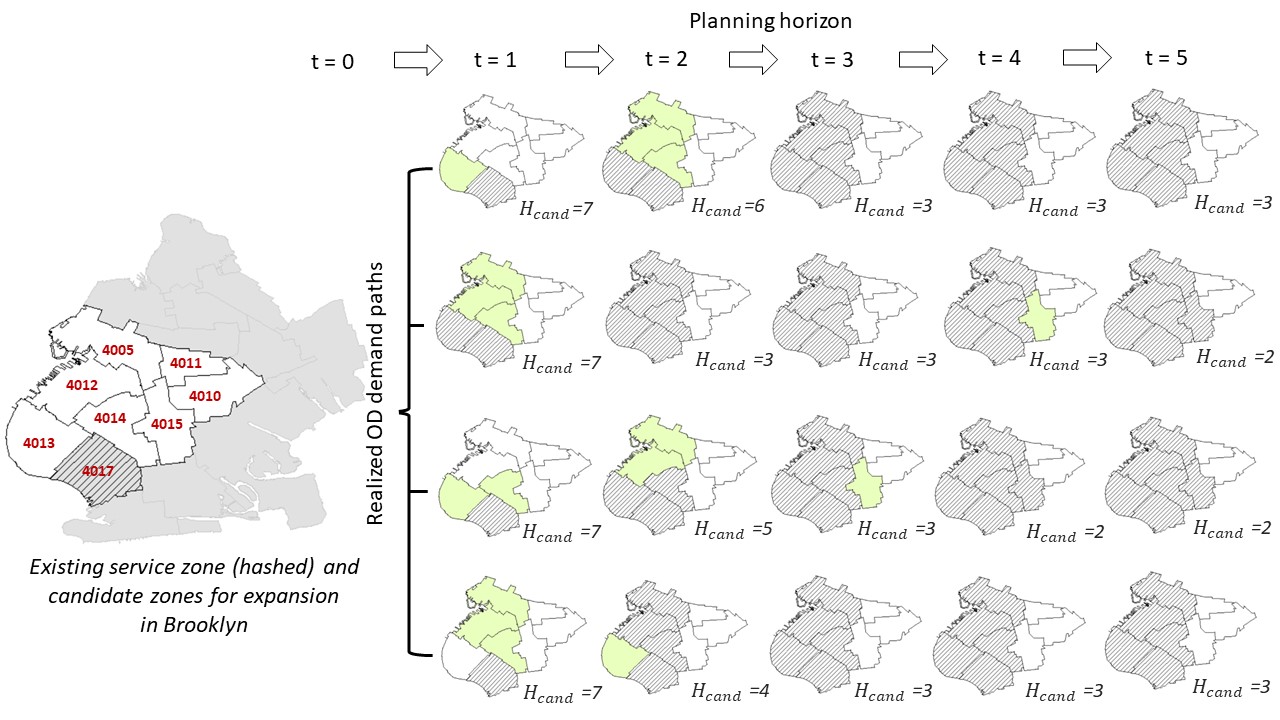}
\caption{CR-RNN policy in the sequential service region design for the expansion of MoD service area in Brooklyn, NYC: figure illustrates zone investment and deferral decisions for four different realized demand paths $\in \mathcal{P}$ in a 5-year rolling time horizon. For each $p-t_e$, already covered zones (in previous time steps) are hashed, zones that are to be invested immediately as per the optimal CR-RNN policy are highlighted in green, and the rest represent zones with investment deferral decisions.} 
\label{fig:case_study} 
\end{center}
\end{figure}

We validate the proposed method in a case study for sequential service region design of MoD services in Brooklyn area (NYC) by comparing its simulated performance to a benchmark policy. Consider an MoD operator with an existing service area
in Bensonhurst neighborhood, Brooklyn (PUMA zone ID 4017) is planning to expand their services to other nearby areas (\emph{i.e.}, 7 candidate PUMA zones, each constituting multiple taxi zones) as illustrated in \Cref{fig:case_study} (numbers on the zones represent PUMA zone IDs). The objective of the operator is to evaluate the added value from the real option strategies to assess alternatives in staging investments for the portfolio of 7 interacting service zones under OD demand uncertainty. Considering a 5-year rolling time horizon, we look at the sequential service region design problem for the 7 service (PUMA) zones, where we evaluate average net present value of the future payoffs based on the optimal investment timing decisions obtained using the CR-RNN policy at different time steps in the planning horizon (as described in \Cref{sec:setup}). 

Assuming a similar setup as \Cref{sec:setup}, we consider the operator has data available for initial (current time) potential OD (taxi zone) demand for MoD services along with zone-wise demand volatility for the $8$ PUMA zones in the region. The potential OD demand values, ridership threshold (\emph{i.e.}, within-zone cost $C_{wz}$ and interzone cost $C_{iz}$), and zone-wise demand volatility ($\sigma_H$) are obtained as per \Cref{sec:experiment_data}a; this gives us $\sigma_H$ = \{4010: 0.3, 4011: 0.15, 4005: 0.3, 4012: 0.15, 4014: 0.25, 4015: 0.05, 4013: 0.35, 4017: 0.1\}, $C_{wz} = 54$, and $C_{iz} = 55$. First, a set of $\mathcal{P}$ independent (GBM) paths of realization are simulated for the stochastic OD demand in the candidate region over the rolling time horizon $\mathcal{E}$ ($|\mathcal{P}| = 10$, $|\mathcal{E}| = 5$). We denote each realization $p \in \mathcal{P}$ during a time step $t_e \in \mathcal{E}$ as $p-t_e$, and refer to it as a decision epoch for simplicity. For each decision epoch, a roll period of $\mathcal{T}= 5$ (future) time periods starting from $t_e$ is considered. 

For a given number of $\mathcal{H}_{cand}$ service zones at a particular $p-t_e$, based on the realized OD demand $D_{p,t_e}$ for the zones, we determine the optimal investment timing decisions for each zone $h \in \mathcal{H}_{cand}$. This is obtained using \Cref{alg:cr_rnn_algo} (with $\mathcal{P}'$ GBM simulated paths based on $D_{p,t_e}$); the parameter values for $\mathcal{P}'$ and other inputs are the same as in Section~\ref{sec:experiment_data}a. Hence, for each $p-t_e$, we find the zones to be invested immediately (as per the CR-RNN policy) and add it to the service region design for a path $p$, while the zones that are deferred are considered for evaluation in the next time step for the path $p$, and so on in a rolling horizon until $t_e$ = $|\mathcal{E}|$. Therefore, at each $p-t_e$, we have a set of already covered service zones ($\mathcal{H}_{cov}$), and the CR-RNN policy evaluates the remaining candidate zones ($\mathcal{H}_{cand}$ = $\mathcal{H}$-$\mathcal{H}_{cov}$) for the service region design. The set of OD pairs for the $\mathcal{H}_{cand}$ zones include the taxi zone pairs across $\mathcal{H}_{cand}$ zones, along with the OD pairs connecting the existing service zones ($\mathcal{H}_{cov}$) and the $\mathcal{H}_{cand}$ zones. Based on this, we use the following modification of \Cref{eq:zone_payoff} for payoff calculation (in Section~\ref{sec:payoff}) for a zone $z_h$ in an ordered set of $\mathcal{H}_{cand}$ zones ($h=\{1,2,3...H_{cand}\}$).
\begin{align}
    \pi^{'}_{z_h} = X_{z_h} - (C_{wz} + 2(h-1+|\mathcal{H}_{cov}|)C_{iz}) \label{payoff_expansion}
\end{align}
where $X_{z_h}$ is the zone ridership. 

\Cref{fig:case_study} shows four sample demand paths for $t_e=\{1,2,3,4,5\}$, and the zone investment decisions as per the optimal CR-RNN policy. For example, consider the first demand path in the figure which shows an already existing service zone (hashed) at $t_e=1$ ($|\mathcal{H}_{cov}|$ = 1, $|\mathcal{H}_{cand}|$ = 7); the optimal CR-RNN policy (on 7! investment sequences) is to invest in one zone (highlighted in green). Adding this to the service region, at $t_e=2$, we determine the optimal investment strategy with $|\mathcal{H}_{cov}|$ = 2 and $|\mathcal{H}_{cand}|$ = 6, and so on. For decision steps $p-t_e$ with $|\mathcal{H}_{cand}| > 6$ zones, we use RNN classifiers to guide the investment decisions, while for $|\mathcal{H}_{cand}| \leq 6$ zones, CR policy (without the RNN component) is used. Following the ML model training process as described in \Cref{sec:CR RNN policy}, we build the RNN classifiers using $6\%$ randomly sampled sequences (from $|H_{cand}|!$ sequences) with $PNR_{max}$ 0.01 to predict top $50$ sequences from the remaining fraction of sequences, which are used to determine the optimal investment strategy for a $p-t_e$.  Details of the zone investment decisions for each $p-t_e$ (\emph{i.e.}, total 50 decision steps for $|\mathcal{E}|=5$ and $|\mathcal{P}|=10$) is included in Appendix~\ref{appendix:casestudy}. For a particular demand path $p$, based on the investment decisions at each time step $t_e \in \mathcal{E}$, we compute the cumulative investment payoffs in terms of ridership from the set of selected service zones at $p-t_e$ (\emph{i.e.}, $|\mathcal{H}_{cov}(t_e,p)| \leq 8$) based on the realized OD demand at that $p-t_e$. The net present value of the future payoffs is calculated using \Cref{eq:npv_futurepayoffs} for each demand path $p \in \mathcal{P}$.
\begin{align}
    NPV_p = \sum^{|E|}_{t_e=1} (1+\rho)^{(-t_e)} \times \sum_{h=1}^{H_{cov}(t_e,p)} \pi_{z_h}(t_e,p) \label{eq:npv_futurepayoffs}
\end{align}
where $\pi_{z,h}(t_e,p)$ denote payoff (\Cref{sec:payoff}), and $\rho$ is the discount rate (see \Cref{sec:experiment_data}a).
The average NPV across all 10 demand paths using the CR policy (\emph{i.e.}, without the proposed RNN component) is 1248 with a total computation time of 87 hours. On the other hand, the average NPV obtained using the CR-RNN policy is 1242 ($\approx$ 0.5\% gap) but with a reduced run time that is about 5.4 times faster.
%from evaluating total 72,334 investment sequences 
%on total 73,030 sequences 
 In most cases, the CR-RNN policy suggests deferring the zones in the eastern part of the candidate region (\emph{i.e}, mainly zone IDs 4010, 4011, 4015) to later time periods in the planning horizon (refer Appendix~\ref{appendix:casestudy}).\\
 
 \paragraph{Comparison with benchmark policy}
 The CR-RNN policy can be compared to different policies for service region design.  In our study, we compare the results from the CR-RNN policy to a benchmark policy which considers investing in all the candidate service zones at the start. In particular, for each decision step $p-t_e$ in the benchmark policy, $\mathcal{H}_{cov}(t_e,p)$ includes all the $8$ service zones in the region. 
Based on the investment decisions obtained using each of the above two policies, the objective is to  evaluate the scope of the MoD operator's profit or investment benefits with respect to the service region design resulting from the respective policies. For this purpose, at each $p-t_e$, we compute the profitability of the investment strategy based on the policy considered. Profitability gives a relative measure of the cumulative investment payoff with respect to the total ridership from the investment at $p-t_e$ based on the selected zones $\mathcal{H}_{cov}(t_e,p)$ (which is essentially a proxy for profit/revenue). The present value of the future profitability measure across $P$ runs is calculated using \Cref{eq:pv_profitability}.
 \begin{align}
    PV_{profit} = \dfrac{1}{|\mathcal{P}|} \times \mathlarger{\sum^{|\mathcal{P}|}_{p=1}\sum^{|\mathcal{E}|}_{t_e=1}} (1+\rho)^{(-t_e)} \times \frac{\sum_{h=1}^{|\mathcal{H}_{cov}(t_e,p)|} \pi_{z_h}(t_e,p)}{\sum_{h=1}^{|\mathcal{H}_{cov}(t_e,p)|} X_{z_h}(t_e,p)} \label{eq:pv_profitability}
\end{align}
%where $\pi_{z,h}(t_e,p)$ and $X_{z_h}(t_e,p)$ denote payoff and ridership respectively (\Cref{sec:CR policy}), and $\rho$ is the discount rate (see \Cref{sec:experiment_data}a).

Using the CR-RNN policy, the average profitability (as per \cref{eq:pv_profitability}) is calculated as 0.44 (95\% CI [0.36, 0.53]), while the value for the benchmark policy is 0.22 (95\% CI [0.12, 0.31]); the average of the difference in $PV_{profit}$ values between the two policies is 0.23 (95\% CI [0.18, 0.28]). We conduct a paired-sample t-test (\citealp{de2013using}) to analyze the mean difference between the two policies, assuming the null hypothesis that the true mean difference is equal to zero. Results show a statistically significant difference between the two policies (with $p-$value $<$ 0.001). This shows there is added value in opting the real option investment strategy for the MoD service region design in the selected region. Similarly, various other alternatives can be considered in the decision process of evaluating the trade-offs between present benefits and deferral option values. In this context, the proposed method can be used to efficiently assess and compare RO investment policy alternatives that can guide MoD operators and planners in strategically and dynamically designing the service region under demand uncertainty.

%% CONCLUSION %%
\section{Conclusion and future works} \label{sec:conclusion}
 This study presents a new ML based framework for efficient valuation of real options strategies for large-scale sequential service region design and timing of MoD services with non-stationary stochastic variables, framed as a Markov decision process. The proposed CR-RNN policy provides quantitative support for flexibility in investment decisions in service region design under time-dependent demand uncertainty associated with such emerging mobility technologies. The proposed model treats each service zone in the region as a separate and interacting option with non-stationary stochastic OD demand flows; the model uses multi-option LSMC simulation to value decision flexibility in determining the optimal investment strategy (investment or deferral decision) for the service zones with heterogeneous demand volatility. For large-scale cases, \emph{i.e.,} for regions with a large number of $H$ compounded zones, the study proposes a supervised ML approach (using an RNN model) to address the high computational cost involved in valuating all possible $H!$ investment sequences via the CR policy (\citealp{chow2011network,chow2016reference}). The objective of the proposed ML method in CR-RNN policy is to accelerate the ROV calculations by selecting only a small set of promising sequences based on learning from randomly sampled sequences from the population set of investment sequences. We frame this as a sequence classification problem, where, for a given set of $H$ service zones and $H!$ sequences, only a small fraction of sampled sequences are used to train an RNN (classifier) model using a sequence labeling strategy (guided by a pre-determined positive to negative sample ratio) to learn the useful patterns in sequences associated with high and low policy values. The trained model is then used to predict the top $K$ promising sequences from the larger fraction of remaining sequences ($K$ being relatively very small compared to the size of the remaining sequence set). 
 
 Based on the experiments conducted on multiple MoD service region scenarios (with 7 and 8 service zones in NYC) with different input parameters, the study demonstrates the effectiveness of the proposed ML approach in predicting the top $K$ promising sequences ($K \approx 50$) with a high accuracy for $PNR_{max}$ in the range 1-2\%, while significantly reducing the overall computational complexity compared against the CR policy without the RNN model (\emph{i.e.}, time reduction associated with the RO evaluation of more than 90\% of total investment sequences is achieved). To illustrate the applicability and feasibility of the proposed CR-RNN policy as a dynamic decision tool, we test the policy in a case study for the sequential service region design of MoD services in NYC Brooklyn area, where the investment decisions are adapted to new simulated (service demand) information in the future. Results from the case study show that using the CR-RNN policy in determining optimal real options investment strategy yields a similar performance ($\approx$ 0.5\% within the CR policy value) with significantly reduced computation time (about 5.4 times faster). For higher number of zones, the proposed method could lead to more time savings with the use of the CR-RNN model at multiple decision epochs. Validation against a benchmark policy also show promising results.
 
 Although the proposed methodology focuses on the service region design problem, it can very well be applied to a wide range of problems with interacting options and with different set of stochastic variables. The use of real options principle in the service region design context gives a lower bound estimate on the true solution (which remains unsolvable). It can support planners by adding a time dimension to the investment decision making process in a dynamically uncertain environment. Future studies should look into incorporating demand and volatility changes in response to the service zone investment decisions, to better reflect reality. A major contribution of the proposed methodology is the use of the RNNs towards learning complex patterns in investment sequences for efficient ROV calculations. To further improve the computational efficiency of the proposed method, one possible extension could be to incorporate active learning techniques (\citealp{settles2009active, zhong2021active}) to effectively curate the training data for the RNN model to achieve similar model performance with relatively smaller yet representative training samples. Exploring new sequence labeling strategies, and the use of other RNN variants (\citealp{deep_learning_book_goodfellow}) or advanced models such as transformers (\citealp{vaswani}) are promising future research directions.
 
This study opens up new possibilities at the intersection of sequential service region design and deep learning based sequence models for building decision support tools, while reinforcing existing studies on the use of ROA in transportation. In addition to addressing the computational complexity associated with timing decisions of a large number of zones, if properly trained (with sufficient number of simulated scenarios), the proposed method can generalize to varying region sizes via fine-tuning a pre-trained ML model with very few samples from the new setup. In this case, the proposed method can be used to efficiently valuate different scenarios for faster training of the generalized ML model with varying region sizes (using zero padding, \citealp{piao2018financial}) along with other region specific input parameters. This can be built into a decision support tool
%to assist planners in efficiently estimating cost and benefits of various resource allocation strategies for a given budget. This can be 
which could benefit both public agencies and mobility operators in the funding and resource allocation planning, as part of a complex cost-benefit analyses process. For example, consider the mobility company Via, which operates microtransit services under different policies in different set of cities, in partnership with their transit agencies. In some cities, the services start with a pilot program covering a small area, which is later on expanded to multiple service zones. While managing a portfolio of cities, the service expansion decisions to new areas or new cities are primarily constrained by available funding or budget along with future uncertainties of service adoption. Moreover, the objective of such service implementations may vary across different cities, and so do the uncertain elements associated with such services. Therefore, based on the purpose and policies of the (existing) services in different cities and the city characteristics, along with the data available to operators, investment payoffs and the stochastic elements in the RO setup can be defined among other inputs to train a generalized ML tool (using simulated scenarios with different objectives, policies, region sizes, etc.). Such a tool can assist planners to easily predict real options investment strategies to assess trade-offs between
present benefits and deferral option values for service region design based on different performance measures or operation policies; this can be used to efficiently compare resource allocation strategies for a portfolio of cities, when considering the  time-dependent uncertainties associated with emerging transportation services.

\section*{Acknowledgement}
\noindent
The authors wish to acknowledge the funding support from C2SMART University Transportation Center (USDOT \#69A3551747124) and National Science Foundation CMMI-1652735.

\bibliography{references} 

\begin{appendices}
\section{LSTM architecture details} \label{appendix:lstm_appendix}
LSTM's proposal came from the need to capture long range dependencies in RNNs (\citealp{deep_learning_book_goodfellow}). To learn such dependencies, an LSTM unit has a memory cell, and has the following gates with sigmoid activations:
\begin{itemize}
    \item input gate to control additions from the current input,
    \item output gate to control control contributions to the next hidden state, and
    \item forget gate to control the the amount of decayed memory.
\end{itemize}
The gates and update rules can be described as shown below:
\begin{align}
    f_t      &= \sigma(W_{fe}^T e_t   +  W_{fd}^T d_{t-1} ), \\
    i_t      &= \sigma( W_{ie}^T e_t  +  W_{id}^T d_{t-1} ), \\
    o_t      &= \sigma( W_{oe}^T e_t  +  W_{od}^T d_{t-1} ), \\
\tilde{c}_t  &= tanh( W_{ed}^T e_t    +  W_{dd}^T d_{t-1} ), \\
\text{cell state update} \; \rightarrow \; c_t          &= f_t \bigodot c_{t-1}  + i_t \bigodot \tilde{c}_t, \\
\text{hidden state update} \; \rightarrow \; d_t          &= o_t \bigodot tanh(c_t),
\end{align}
where $\sigma(\cdot)$ denotes the sigmoid activation function, $d_t$ is the hidden state for LSTM unit $t$,
$c_t$ is the cell state for LSTM unit $t$, $e_t$ is the input to LSTM unit $t$ which is essentially the embedding for the corresponding zone in our setup, $f_t$ is the forget gate activation vector, $i_t$ is the input gate activation vector, $o_t$ is the output gate activation vector, $T$ denotes matrix transpose, and $\bigodot$ is the Hadamard product. The weight matrices $W_{fe}, W_{ie}, W_{oe}, W_{ed}, W_{fd}, W_{id}, W_{od}$, and $W_{dd}$ are parameters which are learnt during training.
The above description captures the standard implementation for an LSTM unit with a forget gate, and additional details can be found in \cite{deep_learning_book_goodfellow}.

\section{CR-RNN policy NYC case study results} \label{appendix:casestudy}
\Cref{tab:case_study_results} shows the optimal RO investment strategies for the $7$ candidate service zones (in the selected region in Brooklyn, NYC) for each realized demand path $p \in P$ during each time step $t_e \in E$ ($|P| = 10, |E| = 5$). These decisions are obtained using the CR-RNN policy (as explained in \Cref{sec:case study NYC}). The table shows the final set of selected service zones (\emph{i.e.}, zones which as per the CR-RNN policy are invested immediately at a certain $t_e$) for each $p$ in the sequential design of the MoD service region; this also includes the already existing service zone (zone ID 4017) in the region.

\begin{table}[h!]
  \centering
  %\begin{tabular}{ | >{\centering\arraybackslash}p{0.8cm} | c | c |c |>{\centering\arraybackslash}p{2cm} | >{\centering\arraybackslash}p{1cm} |>{\centering\arraybackslash}p{2cm} | >{\centering\arraybackslash}p{1cm}| >{\centering\arraybackslash}p{0.7cm}|}
  %\resizebox{\textwidth}{!}{
  %\scalebox{0.99}{
  \begin{tabular}{|c|c|c|c|c|c|}
    \hline
    Demand	&	t = 1	&	t = 2	&	t = 3	&	t = 4	&	t = 5\\ 
   path	($p$) &		&		&	&	&	\\ \hline
    1	&	 4017, 4015, 4013, 	&	 4017, 4015, 4013, 	&	 4017, 4015, 4013, 	&	 4017, 4015, 4013, 	&	 4017, 4015, 4013, \\ 
    	&	  4012, 4005, 4014	&	  4012, 4005, 4014	&	  4012, 4005, 4014	&	  4012, 4005, 4014	&  4012, 4005, 4014 \\ \hline
    2	&	 4017, 4014, 4005,	&	 4017, 4014, 4005,	&	 4017, 4014, 4005, 	&	 4017, 4014, 4005, 	&	 4017, 4014, 4005, \\ 
    	&	  4012	&	  4012, 4013	&	 4012, 4013	&	  4012, 4013	&	  4012, 4013 \\ \hline
    3	&	4017, 4005	&	 4017, 4005	&	 4017, 4005, 4015, 	& 4017, 4005, 4015, 	&	4017, 4005, 4015,  \\ 
    	&		&	 	&	  4012, 4014	&	 4012, 4014	&	 4012, 4014, 4013 \\ \hline
    4	&	 4017, 4012, 4013,	&	 4017,4012, 4013, 	&	 4017, 4012, 4013, 	&	4017, 4012, 4013,  	&	4017, 4012, 4013,  \\ 
    	&	  4005, 4014	&	  4005, 4014	&	  4005, 4014	&	4005, 4014	&	 4005, 4014 \\ \hline
    5	&	 4017, 4005, 4012, 	&	 4017, 4005, 4012, 	&	 4017, 4005, 4012, 	&	4017, 4005, 4012,  	&	4017, 4005, 4012,  \\ 
    	&	4014, 4013	&	 4014, 4013	&	 4014, 4013, 4015	&	4014, 4013, 4015	&	 4014, 4013, 4015 \\ \hline
    6	&	4017	&	4017	&	4017	&	4017	&	4017 \\ \hline
    7	&	 4017, 4012, 4013, 	&	 4017, 4012, 4013, 	&	 4017, 4012, 4013, 	&	4017, 4012, 4013, 	& 4017,	4012, 4013, \\ 
    	&	  4014, 4005	&	  4014, 4005	&	  4014, 4005	&	 4014, 4005	&	 4014, 4005 \\ \hline
    8	&	 4017, 4013, 4014	&	 4017, 4012, 4014, 	&	 4017, 4012, 4014, 	&	4017, 4012, 4014, 	&	4017, 4012, 4014,  \\ 
    	&		&	  4005, 4013	&	  4005, 4013	&	4005, 4013,4015	& 4005, 4013,4015 \\ \hline
    9	&	4017, 4013	&	 4017, 4005,4012, 	&	 4017, 4005, 4012, 	&	4017, 4005,4012, 	&	4017, 4005,4012, \\ 
     	&	&	  4014, 4013	&	 4014, 4013	&  4014, 4013	&	4014, 4013 \\ \hline
    10	&	 4017, 4012, 4014, 	&	 4017, 4012, 4014, 	&	 4017, 4012, 4014, 	&	4017, 4012, 4014, 	&	4017, 4012, 4014, \\ 
     	&	 4013	&	 4013, 4015	&	 4013, 4015, 4005	&	 4013, 4015, 4005	&	 4013, 4015, 4005 \\ \hline
  \end{tabular}%}
  \caption{Sequential service region design using the CR-RNN policy for MoD services in the selected region in Brooklyn, NYC; zone IDs in the table are the NYC PUMA zone IDs}\label{tab:case_study_results}
\end{table}

\end{appendices}

\end{document}